\documentclass{article}

 \usepackage[main, final]{neurips_2026}

\usepackage[utf8]{inputenc} 
\usepackage[T1]{fontenc}    
\usepackage{url}            
\usepackage{booktabs}       
\usepackage{amsfonts}       
\usepackage{nicefrac}       
\usepackage{microtype}      
\usepackage{caption}
\usepackage{wrapfig}
\usepackage{xcolor}         
\definecolor{citecolor}{HTML}{0071bc}
\definecolor{neuripspurple}{HTML}{E7C5F7}

\definecolor{codegreen}{rgb}{0,0.6,0}
\definecolor{negativeorange}{RGB}{214, 39, 40} 
\usepackage{subcaption}
\usepackage{amsmath}

\usepackage{amsthm}
\newtheorem{theorem}{Theorem}
\newtheorem{assumption}{Assumption}

\usepackage[colorlinks, linkcolor=red, colorlinks, anchorcolor=blue, citecolor=citecolor, pagebackref=True]{hyperref}
\usepackage{graphicx}
\usepackage{multirow}
\usepackage[table]{xcolor}

\title{Mixture of Probes: Learning with Privileged Modalities in Multimodal LLMs Through Probing}

\author{
  \textbf{Dominick Reilly}$^{1,2}$\thanks{Work done during internship at Sony Group Corporation} \quad \textbf{Qiyu Wu}$^{1}$\thanks{Corresponding Author: \tt\small {qiyu.wu@sony.com}} \quad \textbf{Hiromi Wakaki}$^{1}$ \quad \textbf{Srijan Das}$^{2}$ \quad \textbf{Yuki Mitsufuji}$^{1}$ \vspace{0.1cm} \\
  $^{1}$ Sony Group Corporation \\ $^{2}$ University of North Carolina at Charlotte \vspace{0.1cm} \\
  {\tt\small dreilly1@charlotte.edu} \\
  {\tt\small \{qiyu.wu, hiromi.wakaki, yuhki.mitsufuji\}@sony.com} \\
  \url{https://github.com/Sony/MoP}
}

\begin{document}

\maketitle

\begin{abstract}
\vspace{-0.25cm}
Multimodal Large Language Models (MLLMs) are typically designed under the assumption that all modalities available during training will also be accessible at inference. However, many real-world settings violate this assumption, requiring models to operate under a \emph{privileged modality setting}, where auxiliary modalities are available only during training. While these modalities contain valuable information, existing MLLMs largely fail to leverage them effectively, as they treat modalities as interchangeable inputs rather than sources of complementary supervision.
We propose \textbf{Mixture of Probes (MoP)}, a novel framework that disentangles \emph{modality-specific} and \emph{modality-general} signals within the MLLM, allowing the model to preserve modality-dependent structure while learning transferable representations across modalities. At its core, MoP achieves this through a structured probing mechanism that extracts and organizes information from intermediate representations of a shared modality encoder, rather than relying only on final-layer alignment as done in existing MLLMs.
To support this disentanglement, we further introduce \textbf{MoP Cross-modal Training (MoP-X)}, a training strategy for MoP centered around a probe disentanglement loss that prevents probe collapse and encourages cross-modal learning.
We evaluate MoP across two domains spanning eight tasks and four modalities under a comprehensive evaluation protocol tailored to the privileged modality setting, where each modality is independently treated as the sole input at inference time. MoP consistently outperforms strong MLLM baselines, achieving up to 65\% relative improvement, demonstrating that auxiliary modalities, even when unavailable at inference, can provide substantial gains when effectively leveraged during training.
\vspace{-0.5cm}
\end{abstract}

\section{Introduction}
\vspace{-0.25cm}
Stemming from the success of large language models (LLMs)~\cite{touvron2023llama, brown2020gpt3, gemmateam2024gemma}, recent multimodal large language models (MLLMs) have enabled language-based interaction with diverse sensory inputs such as vision, audio, and depth~\cite{zhang2023videollama, han2024imagebindllm, openai2024gpt4ocard, xu2025qwenomni25}. Yet real-world scenarios often exhibit an asymmetric structure: the sensors that are available for training are not always available for deployment. A model may be trained with rich auxiliary streams, such as egocentric video, depth, or audio, but later be required to operate from a single cheap, reliable, or non-invasive modality. This creates a \emph{privileged modality setting}, where auxiliary modalities provide supervision during training but are unavailable at inference. For example, in activities of daily living (ADL), training data may include egocentric video from wearable devices~\cite{cvpr2025egoexo4d, ego4d}, depth from specialized sensors~\cite{shotton2013kinect}, and exocentric video from static cameras~\cite{das2019toyotasmarthome, dai2022toyotasmarthomeuntrimmed}, while deployment is typically restricted to exocentric video alone.

Existing MLLMs~\cite{videollama, videollava, timechat, Qwen2.5-VL} are not designed to address this setting. They primarily view them as interchangeable inputs rather than treating modalities as sources of complementary supervision, exclusively aiming only to align each modality into a shared representation space~\cite{panagopoulou2024xinstructblip, zhao2023chatbridge, moon2024anymal, han2024onellm}. 
However, in the privileged modality setting, it is not sufficient for a model to merely align each modality; it must leverage information from modalities available only during training and transfer this knowledge to improve performance on the modality available at inference.
Most methods achieve such alignment with modality-specific encoders~\cite{panagopoulou2024xinstructblip, zhao2023chatbridge, moon2024anymal}, or through the more scalable unified modality encoder paradigm~\cite{han2024onellm}.
We identify two key limitations of these MLLMs in the privileged modality setting:
(i) \textbf{sub-optimal modality representations} - new modalities are typically incorporated via shared encoders with limited modality-specific adaptation at the final layers, neglecting intermediate features that capture low-level modality-specific signals~\cite{raghu2021vitseelikecnn};
(ii) \textbf{unstructured cross-modal representations} - existing approaches lack explicit mechanisms to disentangle modality-general and modality-specific information, hindering effective cross-modal knowledge transfer~\cite{xue2023modality}.

\begin{figure}[t]
    \centering
    \includegraphics[width=\linewidth]{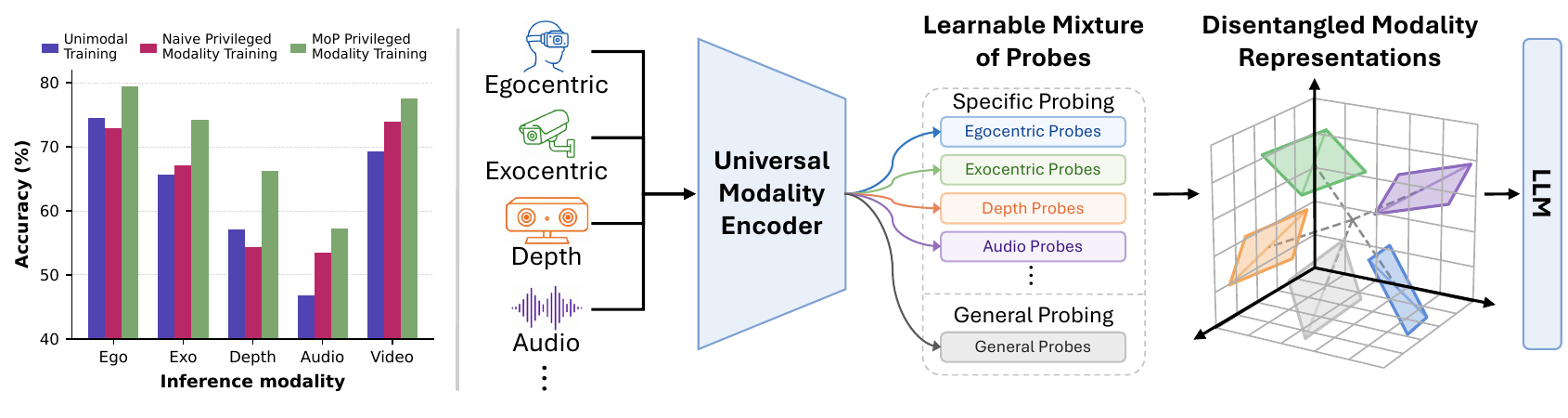}
    \vspace{-0.7cm}
    \caption{\textbf{(left)} MoP improves single-modality inference across five inference settings, outperforming both unimodal MLLMs and naive multimodal training. \textbf{(right)} MoP achieves this through a structured probing mechanism with modality-specific and modality-general probes, which disentangle modality representations before integration with the LLM.}
    \label{fig:teaser}
    \vspace{-0.5cm}
\end{figure}

To this end, we propose \textbf{Mixture of Probes (MoP)}, an MLLM architecture for the privileged modality setting that improves cross-modal knowledge transfer by encouraging disentangled \emph{modality-specific} and \emph{modality-general} representations to be learned through a structured probing mechanism.
MoP treats all modalities symmetrically and designates none as privileged; privilege arises solely from which modality is available at deployment.
Inspired by evidence from neuroscience that the human brain separates modality-specific maps from modality-general abstractions across sensory streams~\cite{dang2024neuro_specific-general}, MoP is built around the idea that effective cross-modal transfer requires preserving what is unique to each modality while isolating what can be shared across them. This intuition is further supported by the Modality Focusing Hypothesis~\cite{xue2023modality}, which finds that cross-modal transfer depends primarily on modality-general decisive features rather than modality-specific signals that do not transfer across modalities.
Concretely, MoP introduces a structured probing mechanism that leverages intermediate representations throughout a shared universal encoder, enabling richer extraction of both modality-dependent and shared signals across the model depth. Modality-specific probes are conditioned only on individual modalities to capture modality-specific structure, while modality-general probes are conditioned on all modalities to capture general cross-modal structure, illustrated in Figure \ref{fig:teaser}.
A key component of our design is \textbf{MoP Cross-modal Training (MoP-X)}, which consists of two components: (i) a \emph{probe disentanglement loss} that plays a central role by explicitly encouraging disentanglement between the modality-specific and modality-general probes and prevents probe collapse; (ii) \emph{modality-interleaved batching} that ensures consistent cross-modal interaction during training, unlike prior methods that train each modality sequentially or in isolation~\cite{panagopoulou2024xinstructblip, zhao2023chatbridge}.

We evaluate MoP on \textbf{two domains} spanning \textbf{8 tasks} and \textbf{4 modalities}. In the \textit{Activities of Daily Living} (ADL) domain, MoP is trained using egocentric video, exocentric video, and depth. In the \textit{Music Understanding} domain, MoP is trained on paired audio-visual data. In each domain, we perform a comprehensive evaluation under the privileged modality setting, where each modality is independently treated as the sole input at inference time. Across both domains and all inference conditions, MoP consistently outperforms state-of-the-art MLLM baselines, demonstrating that auxiliary modalities, when properly leveraged, can substantially improve single-modality performance.


\noindent A summary of our contributions follows:
\vspace{-0.2cm}
\begin{itemize}
    \itemsep0em
    \item We formalize the \textbf{privileged modality setting} for MLLMs, where auxiliary modalities enrich training but are absent at inference, and demonstrate that standard multimodal training fails to effectively leverage this asymmetry.

    \item We propose \textbf{Mixture of Probes (MoP)}, a novel probing-based architecture that learns disentangled modality-specific and modality-general representations through learnable probing across the depth of a modality encoder.
    
    \item We introduce \textbf{MoP Cross-modal Training (MoP-X)}, a training strategy complementing MoP via a probe disentanglement loss to separate modality-specific from modality-general representations, and promote cross-modal interaction through modality-interleaved batching.
    
    \item We demonstrate the effectiveness of MoP across two domains, eight tasks, and four modalities under the privileged modality setting, consistently outperforming strong MLLM baselines and highlighting the benefits of leveraging privileged modalities.
\end{itemize}

\vspace{-0.5cm}
\section{Related Works}
\vspace{-0.25cm}
\noindent\textbf{Multimodal Large Language Models.}
Multimodal Large Language Models (MLLMs) extend LLMs beyond text by mapping sensory inputs (e.g., vision, video) into a shared language representation space~\cite{liu2023_llava, beyer2024paligemma, dai2023instructblip, maaz2024videochatgpt, ye2024mplugowl2}. To support multiple modalities, many approaches introduce specific modality encoders and alignment modules to connect each modality to the LLM~\cite{zhang2023videollama,chen2023xllm,zhao2023chatbridge,moon2024anymal,su2023pandagpt,han2024imagebindllm,panagopoulou2024xinstructblip, reilly2025llavidal}, while others such as OneLLM use a shared universal encoder across modalities~\cite{han2024onellm}, motivated by previous works demonstrating the robustness of transformer representations~\cite{lu2022pretrainedtransformer,zhang2023metatransformer}. However, these methods are primarily designed for \emph{modality alignment}, treating each modality as an input to be independently mapped into the LLM space. In contrast, we study the \emph{privileged-modality} setting, where auxiliary modalities are available only during training and must improve a single-modality at inference, requiring explicit cross-modal knowledge transfer rather than alignment alone.

\noindent\textbf{Learnable Token-based Adaptation Mechanisms.}
A related line of work adapts pretrained vision-language models using small sets of learnable tokens that interface with frozen backbones. Methods such as BLIP-2~\cite{blip2} and InstructBLIP~\cite{dai2023instructblip} introduce query tokens to extract task-relevant information from a vision encoder and align it with an LLM, while prompt tuning approaches adapt transformers through learnable tokens in the input without modifying the backbone~\cite{jia2022visualprompttuning}.
However, these approaches treat learned tokens as a shallow interface to the model, typically interacting only with final encoder representations and capturing a single undifferentiated signal. In contrast, our approach uses probing as a mechanism for \emph{structured representation learning}, inserting probes throughout the encoder to extract information at different levels of abstraction and explicitly disentangling modality-general and modality-specific signals. This enables the model to not only adapt, but to organize information to support effective cross-modal transfer in the privileged-modality setting.

\noindent\textbf{Learning with Missing or Privileged Modalities.}
Learning with missing or privileged modalities has been widely studied across multimodal learning, domain adaptation, and knowledge distillation. The classical Learning Using Privileged Information (LUPI) framework~\cite{vapnik2009lupi} introduces auxiliary modalities available only at training time, later extended to deep learning via generalized distillation~\cite{lopez2016generalized}. In multimodal settings, related ideas appear in cross-modal distillation and supervision, where one modality guides another (e.g., RGB $\rightarrow$ depth, audio $\rightarrow$ vision)~\cite{gupta2016crossmodal,aytar2016soundnet,owens2016ambient}. More recent work addresses the \emph{missing modality} setting, focusing on robustness when inputs are incomplete through modality dropout, reconstruction, or invariant fusion strategies~\cite{neverova2016moddrop,ma2021missing,wu2026incomplete}, with emerging extensions to multimodal LLMs that compensate for missing inputs via retrieval or tool use~\cite{pipoli2025missrag}.

Despite these advances, existing approaches primarily emphasize robustness to missing modalities or reconstruction of absent inputs, rather than explicitly leveraging auxiliary modalities to improve a target modality at inference. Moreover, existing methods are developed for task-specific multimodal models, making them non-trivial to adapt to MLLMs, where modalities must be integrated through a shared language-centered representation space. In contrast, we focus on \emph{privileged-modality transfer} for MLLMs, where auxiliary modalities are available only during training and are used to directly improve single-modality performance at inference.

\vspace{-0.3cm}
\section{Problem Formulation}
\label{sec:problem_formulation}
\vspace{-0.25cm}

Let $\mathcal{M} = \{1, \dots, M\}$ denote the set of modalities available during training. Each sample in our training dataset consists of multi-modal observations of the same scene
\setlength{\abovedisplayskip}{3pt}
\setlength{\belowdisplayskip}{3pt}
\begin{equation}
    \mathbf{X} = \{X^{m}\}_{m \in \mathcal{M}},
\end{equation}
where $X^{m}$ denotes the input from modality $m$ (e.g., egocentric video, exocentric video, depth, or audio), together with a language query $q$ and a target sequence $y = (y_1, \dots, y_T)$.

We consider the \emph{privileged modality} setting, in which all modalities are accessible during training, but only a \textit{single} modality $X^{(m^\star)}$, where $m^\star \in \mathcal{M}$, is available at test time. 
Privilege is a property of the training and deployment protocol rather than of any particular modality: $m*$ is unknown during training, and whichever modalities are absent at inference act as privileged relative to $m*$.
We aim to learn a predictor
\begin{equation}
    f: (X^{(m^\star)}, q) \mapsto y,
\end{equation}
that benefits from additional modalities during training despite their absence at inference.
Formally, we aim to leverage the full multi-modal training distribution to learn representations or inductive biases that improve generalization under single-modality inference. This differs from standard multi-modal learning, which assumes $\{X^m\}_{m \in \mathcal{M}}$ are available at both training and test time, and instead aligns with the paradigm of learning with privileged information, where auxiliary modalities act as training-time-only supervision.

\vspace{-0.3cm}
\section{Method}
\vspace{-0.25cm}
\subsection{Preliminaries}
\label{sec:preliminaries}
\vspace{-0.15cm}
Recent MLLMs increasingly adopt a \textit{unified architecture}, in which a universal encoder maps heterogeneous input modalities into a shared representation space, rather than relying on separate encoders for each modality. In this paradigm, lightweight modality-specific tokenizers first convert raw inputs into token sequences, which are processed by a shared universal modality encoder. The resulting features are projected into the language embedding space and fed to an LLM together with a text instruction. Below, we describe each component of this unified MLLM framework in more detail.

\noindent\textbf{Modality-specific tokenizers.}
For an input observation $X^{m}$ from modality $m \in \mathcal{M}$, a modality-specific tokenizer $\tau^{m}(\cdot)$ first converts the raw input into a sequence of modality tokens
\begin{equation}
    \tilde{Z}^{m} = \tau^{m}(X^{m}) \in \mathbb{R}^{N^{m} \times d_v}
\end{equation}
where $N^{m}$ is the number of input tokens and $d_v$ is the hidden dimension of the Universal modality encoder. These tokenizers account for modality-dependent input structure while mapping all modalities into a universal token embedding space.

\noindent\textbf{Universal modality encoder.}
All tokenized modality inputs are processed by a shared universal encoder $E$ composed of $L$ Transformer layers. Given the tokenized input $\tilde{Z}^{m}$ for modality $m$, the encoder produces a sequence of hidden representations
\begin{equation}
    H^{m}_l = E_l(H^{m}_{l-1}), \qquad l = 1, \dots, L,
\end{equation}
with $H^{m}_0 = \tilde{Z}^{m}$ and $H^{m}_l \in \mathbb{R}^{N^{m} \times d_v}$. 
Encoder parameters are shared across all modalities, enabling inputs from heterogeneous sensor streams to be mapped into a common latent representation space.

\noindent\textbf{Universal modality connector.}
The final-layer encoder representation $H^{m}_L$ is projected into the LLM embedding space through a universal modality connector $g(\cdot)$:
\begin{equation}
    Z^{m} = g(H^{m}_L) \in \mathbb{R}^{N^{m} \times d_{\mathrm{lm}}}
\end{equation}
where $d_{\mathrm{lm}}$ is the hidden dimension of the LLM. The connector aligns encoder features with the token embedding space of the language model, enabling inputs from different modalities to be consumed through a unified interface.

\noindent\textbf{LLM.}
The projected modality tokens $Z^{m}$ are concatenated with the text tokens  $q$ and provided to the LLM for autoregressive generation. In this way, the LLM operates over a unified sequence consisting of modality-conditioned visual tokens and textual context, producing the output response in the standard causal decoding manner. 

However, such unified MLLMs exhibit several limitations.
\textbf{First,} the universal modality encoder is typically initialized from a pretrained unimodal model, kept frozen, and adapted to new modalities using only its final-layer representations, ignoring useful modality-dependent signals present in lower-level features.
\textbf{Second,} they do not explicitly disentangle modality-general and modality-specific information, which restricts their ability to capture complementary signals and transfer knowledge across modalities~\cite{xue2023modality}.
\textbf{Third,} existing training procedures do not explicitly encourage the learning of complementary information across modalities, as modalities are often treated independently or without structured interactions during training.

\begin{figure}[th!]
    \centering
    \includegraphics[width=0.96\linewidth]{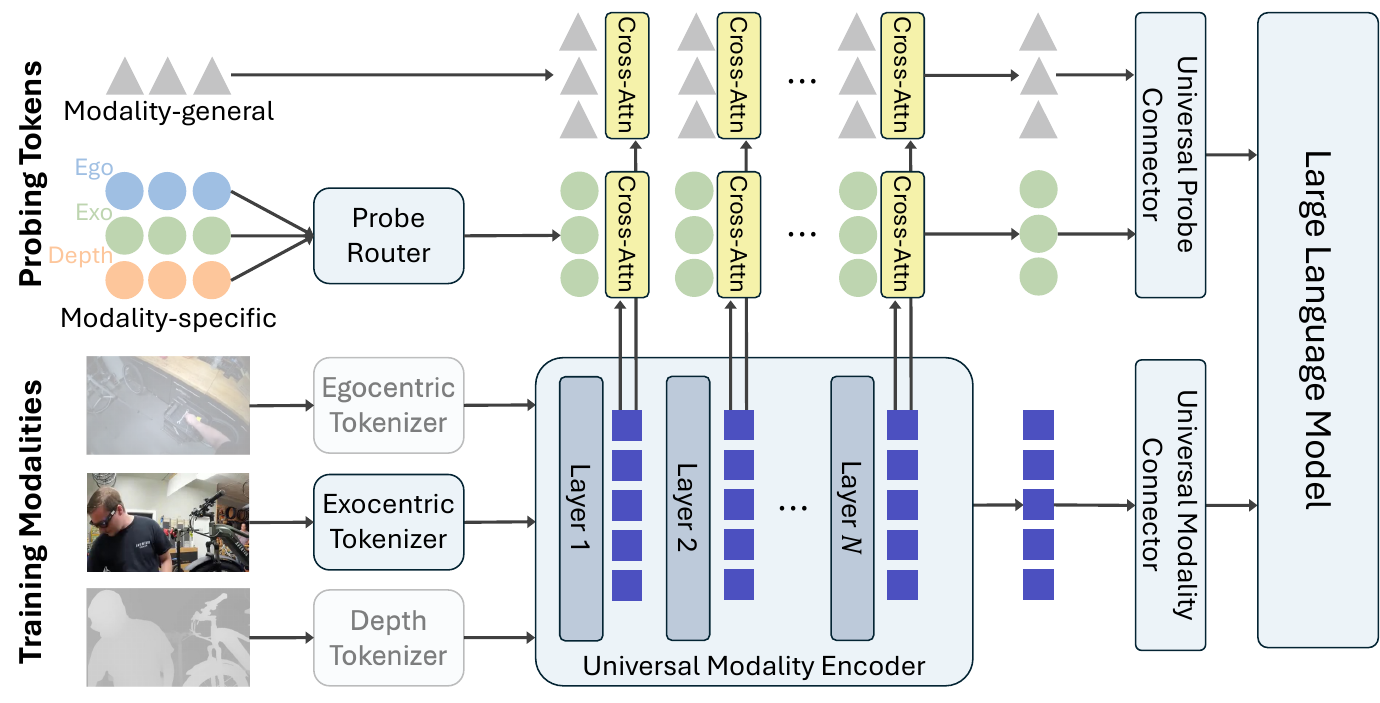}
    \caption{\textbf{MoP Architecture.} In this example, three modalities are shown: egocentric, exocentric, and depth. Each sample in a training batch consists of a single modality, in this example that modality is the exocentric video.}
    \label{fig:method_diagram}
    \vspace{-0.5cm}
\end{figure}

\vspace{-0.15cm}
\subsection{Mixture of Probes (MoP)}
\label{sec:mop_models}
\vspace{-0.15cm}
We introduce \emph{Mixture of Probes (MoP)}, a probing-based architecture that augments a universal modality encoder with learnable tokens operating on intermediate representations across layers. By probing the encoder beyond its final-layer features, MoP extracts salient cues that may not be preserved in the final representation.
Motivated by the modality focusing hypothesis~\cite{xue2023modality}, MoP learns two complementary probe families: modality-specific probes, which capture signals unique to each modality, and modality-general probes, which capture shared structure across modalities.
This design allows the model to represent both shared and modality-specific information within a unified encoder to facilitate effective cross-modal representation learning. We illustrate the MoP architecture in Figure \ref{fig:method_diagram} and present each component in more detail below.

\noindent\textbf{Modality-specific and modality-general probes.}
For each modality $m \in \mathcal{M}$, we define a set of modality-specific probes  $P_s^m \in \mathbb{R}^{K_s \times d_v}$
where $K_s$ is the number of modality-specific probes. In addition, we define a shared set of modality-general probes  $P_g \in \mathbb{R}^{K_g \times d_v}$ where $K_g$ is the number of modality-general probes. Both probe sets lie in the same hidden space as the universal modality encoder and are implemented as a set of learnable tokens.

\noindent\textbf{Layer-wise probe interaction.}
Given tokenized modality input $\tilde{Z}^m$, a deterministic probe router selects the corresponding modality-specific probes for modality $m$, which are passed along with the modality-general probes through every layer of the universal modality encoder and updated through cross-attention with the encoder hidden states. Let $P_{s,l}^m$ and $P_{g,l}$ denote the modality-specific and modality-general probes after interaction with encoder layer $l$, respectively. We initialize $P_{s,0}^m = P_s^m$ and $P_{g,0} = P_g$.
Then, for each layer $l = 1, \dots, L$, the probes are updated as
\begin{equation}
    P_{s,l}^m = \mathrm{CrossAttn}_{s}^{m,l}(P_{s,l-1}^m, H_l^m); \qquad P_{g,l} = \mathrm{CrossAttn}_{g}^{l}(P_{g,l-1}, H_l^m)
\end{equation}
where the probe tokens act as queries and the encoder hidden states $H_l^m$ act as keys and values. Here, $\mathrm{CrossAttn}_{s}^{m,l}$ denotes the modality-specific cross-attention module for modality $m$ at layer $l$, while $\mathrm{CrossAttn}_{g}^{l}$ denotes the corresponding cross-attention module used to update the shared modality-general probes. In this way, the probes iteratively extract modality-relevant information from the universal encoder's intermediate representations across depth.

Importantly, modality-general probes are shared across modalities but interact with one modality per sample, so cross-modal transfer is achieved indirectly through shared parameter updates across modality-interleaved training iterations. This design is intentional, as performing interaction between jointly would substantially increase the token budget and exceed the context length of the MLLM.

\noindent\textbf{Probe-conditioned language tokens.}
After the final encoder layer $L$, we obtain the final modality-specific and modality-general probe representations $P_{s,L}^m  \in \mathbb{R}^{K_s \times d_v}$ and $P_{g,L} \in \mathbb{R}^{K_g \times d_v}$.
The probe tokens are concatenated to form the full probe representation $P^m$, and then projected into the LLM embedding space through a shared probe connector $g_p(\cdot)$:
\begin{equation}
    Z_p^m = g_p(P^m) \in \mathbb{R}^{(K_g + K_s) \times d_{\mathrm{lm}}}
\end{equation}
The resulting probe tokens are then concatenated with the projected universal encoder tokens to obtain $\hat{Z}^m = [Z^m ; Z_p^m]$, and fed to the LLM.
Thus, the LLM receives both the backbone modality tokens and the additional probe tokens that encode modality-specific and modality-general signals.

\vspace{-0.15cm}
\subsection{MoP Cross-modal Training (MoP-X)}
\label{sec:mop_training}
\vspace{-0.15cm}
We train MoP to encourage cross-modal representation transfer in the privileged modality setting. While the probing architecture of MoP exposes both modality-specific and modality-general signals, effective transfer requires these representations to remain distinct. MoP-X enforces this through a probe disentanglement loss, along with a simple modality-interleaved batching strategy to ensure consistent cross-modal interaction during training. During training, we optimize the modality-specific tokenizers, the modality-language connector, the MoP components, and the LLM through LoRA, while keeping the universal modality encoder frozen.

\noindent\textbf{Probe Disentanglement Loss.}
Although the modality-specific and modality-general probes are designed to capture complementary information, this separation is not guaranteed by architecture alone and in practice, probes may collapse toward similar representations. To mitigate this, we introduce a probe disentanglement loss that penalizes inter and intra-probe similarity within each encoder layer. 
Formally, let
\begin{equation}
    P_l^m = [P_{g,l} \,;\, P_{s,l}^m] \in \mathbb{R}^{K \times d_v},
\end{equation}
denote the concatenated modality-general and modality-specific probes at layer $l$, where $K = K_g + K_s$. We perform normalization along the feature dimension and compute the cosine similarity:
\begin{equation}
    S_l^m = \bar{P}_l^m (\bar{P}_l^m)^\top \in \mathbb{R}^{K \times K},
\end{equation}
where $\bar{P}_l^m$ denotes the row-wise $\ell_2$ normalized probe matrix.
We then encourage \textit{only} the off-diagonal entries of $S_l^m$. Let $D \in \mathbb{R}^{K \times K}$ denote the matrix with zeros on the diagonal and ones off-diagonal, the probe probe disentanglement loss is computed by averaging the layer-wise loss across layers:
\begin{equation}
    \mathcal{L}_{\mathrm{dis}} 
    = 
    \frac{1}{L K (K-1)} 
    \sum_{l=1}^{L}
    \left\| 
    D \odot S_l^m 
    \right\|_1,
\end{equation}

\noindent\textbf{Modality-interleaved Batching Strategy.}
We construct mini-batches by sampling examples across multiple modalities, as opposed to previous methods that process each modality sequentially or in isolation~\cite{zhao2023chatbridge, panagopoulou2024xinstructblip}. This exposes shared model components to heterogeneous modality signals within a single optimization step.
While each training sample contains a single modality, interaction between modality-specific and modality-general components is facilitated through shared representations, rather than explicit cross-modal coupling at the sample level.
We favor indirect transfer through shared components over explicit paired objectives for two reasons: jointly aligning time-synchronized modalities induces conflicting gradients~\cite{reilly2025llavidal}, and single-modality samples match the inference distribution while avoiding a multi-modality forward pass per sample.

\noindent\textbf{Autoregressive Loss.}
Given an input from modality $m \in \mathcal{M}$, let $\hat{Z}^m$ denote the final sequence of projected encoder and probe tokens produced by MoP, as defined in Section~\ref{sec:mop_models}. Conditioned on $\hat{Z}^m$ and $q$, the model is trained with the standard autoregressive objective
\begin{equation}
    \mathcal{L}_{\mathrm{AR}}
    =
    - \sum_{t=1}^{T}
    \log p_\theta \!\left( y_t \mid \hat{Z}^m, q, y_{<t} \right)
\end{equation}
where $\theta$ denotes the trainable parameters of the model and $y_{<t}$ represents the subsequence of tokens preceding position $t$. The final loss used for training is the $\lambda$-weighted combination of the probe disentanglement loss and autoregressive loss:
\begin{equation}
    \mathcal{L} = \mathcal{L}_{\mathrm{AR}} + \lambda\mathcal{L}_{\mathrm{dis}}
\end{equation}

\vspace{-0.25cm}
\section{Experiments}
\vspace{-0.25cm}
\subsection{Experimental Setup}
\vspace{-0.15cm}
\noindent\textbf{Benchmarks} We evaluate our approach on two distinct task domains: Activities of Daily Living (ADL) understanding and Music Understanding. For ADL understanding, we evaluate on \textbf{Ego-in-Exo Perception}~\cite{reilly2026egoexo}, a benchmark containing 3,881 questions spanning four task categories, where each question corresponds to a time-synchronzied ego-exo video pair. We evaluate models using a single input modality per question, allowing us to assess performance under different inference sensors. 
Specifically, we consider egocentric video, exocentric video, and depth (obtained from exocentric video using DepthAnythingV2~\cite{yang2024depthanythingv2}) as separate evaluation settings.

For Music Understanding, we evaluate on \textbf{Music-AVQA}~\cite{li2022musicavqa}, which contains 9,185 questions across five music understanding tasks.
Although each question is paired with audio and video modalities, we evaluate each question using a single input modality per sample to stay consistent with our problem setting. We report results under both audio and video inference, allowing controlled analysis of how information from privileged modalities at training time transfers to different inference modalities.

\noindent\textbf{Compared Models.}
We compare MoP against three baseline settings. 
(i) \textbf{Unimodal MLLM}, trained using only the target inference modality without access to privileged modalities. 
(ii) \textbf{Naive MLLM}, 
trained on the same modality-interleaved data stream as MoP, where each sample contains a single modality with its instruction, but without probing or the disentanglement loss.
(iii) \textbf{Existing MLLMs}, including X-InstructBLIP~\cite{panagopoulou2024xinstructblip} and OneLLM~\cite{han2024onellm}, which represent state-of-the-art approaches to multimodal alignment. 
All methods leveraging privileged modalities share the same training data, differing only in how cross-modal information is modeled.

\noindent\textbf{MoP Implementation Details}
We instantiate the universal modality encoder as SigLIP~\cite{zhai2023siglipv1} and the language model as Qwen 2.5~\cite{qwen2025qwen25technicalreport}, both initialized from large-scale pretrained weights~\cite{damonlpsg2025videollama3}. We train on instruction pairs from the EgoExo4D dataset~\cite{reilly2026egoexo} for ADL understanding, and the training split of Music-AVQA for music understanding. Models are trained for 1 epoch on 4 NVIDIA H200 GPUs.

During training, each sample is randomly assigned a single input modality, resulting in mixed-modality batches that expose the model to diverse modalities within each optimization step (Section \ref{sec:mop_training}). The universal modality encoder is kept frozen and we train the modality-specific tokenizers, the universal modality-language connector, and the MoP components, while adapting the language model with LoRA of rank $r=16$. We use a learning rate of $10^{-5}$ for the universal connectors and LoRA parameters, and $2 \times 10^{-6}$ for the modality-specific tokenizers and probe cross-attention modules. Unless otherwise specified, we use 8 modality-specific probes and 8 modality-general probes. We initialize the probes from $\mathcal{N}(0, 0.02)$, and the probe disentanglement loss weight is set to $\lambda = 100$.

\begin{table*}[t!]
    \setlength{\tabcolsep}{5pt}
    \renewcommand{\arraystretch}{1.2}
    \centering
    \caption{\textbf{Results on Ego-in-Exo Perception.} Results are reported on the four tasks of Ego-in-Exo Perception under single modality inference with egocentric, exocentric, or depth modalities. Tasks are abbreviated as: Action Und. = \textit{Action Understanding}, Task Regions = \textit{Task Relevant Regions}, HOI = \textit{Hand Object Interactions}, and Hand Ident. = \textit{Hand Identification}.}
    \vspace{-0.2cm}
    \resizebox{0.98\linewidth}{!}{
    \begin{tabular}{l|ccccc|ccccc|ccccc}
         \hline
         \multirow{3}{*}{\textbf{Method}}&
         \multicolumn{5}{c|}{\cellcolor{gray!20}\textbf{Egocentric Inference}} &
         \multicolumn{5}{c|}{\cellcolor{gray!20}\textbf{Exocentric Inference}} &
         \multicolumn{5}{c}{\cellcolor{gray!20}\textbf{Depth Inference}} \\
         & \shortstack{\textbf{Action}\\\textbf{Und.}} & \shortstack{\textbf{Task}\\\textbf{Regions}} & \textbf{HOI} & \shortstack{\textbf{Hand}\\\textbf{Ident.}} & \textbf{Avg}
         & \shortstack{\textbf{Action}\\\textbf{Und.}} & \shortstack{\textbf{Task}\\\textbf{Regions}} & \textbf{HOI} & \shortstack{\textbf{Hand}\\\textbf{Ident.}} & \textbf{Avg}
         & \shortstack{\textbf{Action}\\\textbf{Und.}} & \shortstack{\textbf{Task}\\\textbf{Regions}} & \textbf{HOI} & \shortstack{\textbf{Hand}\\\textbf{Ident.}} & \textbf{Avg} \\
         \hline
         \multicolumn{16}{c}{\cellcolor{gray!20}\textit{Trained without privileged modalities}} \\
         \hline
         Unimodal MLLM~\cite{damonlpsg2025videollama3}   & 77.04 & 80.49 & 76.03 & 64.74 & 74.57 & 59.50 & 66.60 & 71.60 & 65.00 & 65.68 & 44.89 & 59.88 & 59.74 & 63.71 & 57.06 \\
        \hline
        \multicolumn{16}{c}{\cellcolor{gray!20}\textit{Trained with privileged modalities (Ego+Exo+Depth)}} \\
        \hline
        X-InstructBLIP~\cite{panagopoulou2024xinstructblip} & 77.31 & 76.22 & 76.15 & 65.38 & 73.77 & 61.31 & 68.54 & 71.31 & 65.12 & 66.57 & 46.90 & 50.85 & 62.34 & 65.12 & 56.30 \\
        OneLLM~\cite{han2024onellm} & 35.14 & 35.73 & 47.93 & 63.71 & 45.63 & 23.17 & 35.73 & 46.28 & 62.16 & 41.84 & 21.02 & 32.44 & 44.27 & 62.29 & 40.01 \\
        Naive MLLM~\cite{damonlpsg2025videollama3}       & 75.30 & 75.98 & 75.44 & 65.12 & 72.96 & 60.68 & 69.39 & 73.08 & 64.99 & 67.04 & 41.06 & 52.20 & 59.74 & 64.48 & 54.37 \\
        \rowcolor{neuripspurple}\textbf{MoP (Ours)}       & 84.48 & 86.95 & 80.76 & 65.64 & \textbf{79.46} & 82.68 & 71.68 & 76.86 & 65.51 & \textbf{74.18} & 56.23 & 76.10 & 67.53 & 64.99 & \textbf{66.21} \\
        \hline
    \end{tabular}}
    \label{tab:sota_adl-understanding}
    \vspace{-0.5cm}
\end{table*}

\begin{table*}[h!]
    \setlength{\tabcolsep}{6pt}
    \renewcommand{\arraystretch}{1.2}
    \centering
    \vspace{-0.2cm}
    \caption{\textbf{Results on Music-AVQA.} Results are reported under single-modality inference. Audio-answerable questions are evaluated using audio alone, and video-answerable questions are evaluated using video alone, following the task definitions of Li et al.~\cite{li2022musicavqa}.}
    \vspace{-0.2cm}
    \resizebox{0.96\linewidth}{!}{
    \begin{tabular}{l|ccc|ccc}
         \hline
         \multirow{2}{*}{\textbf{Method}}&
         \multicolumn{3}{c|}{\cellcolor{gray!20}\textbf{Audio Inference}} &
         \multicolumn{3}{c}{\cellcolor{gray!20}\textbf{Video Inference}} \\
         & \textbf{Counting (A)} & \textbf{Comparative (A)} & \textbf{Avg}
         & \textbf{Counting (V)} & \textbf{Location (V)} & \textbf{Avg} \\
         \hline
         \multicolumn{7}{c}{\cellcolor{gray!20}\textit{Trained without privileged modalities}} \\
         \hline
         Unimodal MLLM~\cite{damonlpsg2025videollama3} & 44.84 & 48.65 & 46.75 & 80.40 & 58.22 & 69.31 \\
         \hline
         \multicolumn{7}{c}{\cellcolor{gray!20}\textit{Trained with privileged modalities (Audio+Video)}} \\
         \hline
         X-InstructBLIP~\cite{panagopoulou2024xinstructblip} & 42.18 & 52.05 & 47.12 & 45.12 & 59.84 & 52.48 \\
         OneLLM~\cite{han2024onellm} & 59.98 & 47.47 & 53.73 & 65.58 & 69.71 & 67.65 \\
         Naive MLLM~\cite{damonlpsg2025videollama3} & 53.13 & 53.74 & 53.44 & 83.49 & 64.25 & 73.87 \\
         \rowcolor{neuripspurple}\textbf{MoP (Ours)} & 62.05 & 52.36 & \textbf{57.21} & 83.90 & 71.25 & \textbf{77.58} \\
         \hline
    \end{tabular}}
    \vspace{-0.4cm}
    \label{tab:sota_music-understanding}
\end{table*}

\subsection{Main Results}
\label{sec:experiments_main_results}
\vspace{-0.2cm}
\noindent\textbf{Ego-in-Exo Perception.}
Table \ref{tab:sota_adl-understanding} reports results on Ego-in-Exo Perception across egocentric, exocentric, and depth inference. We observe that unimodal baselines consistently outperform the Naive MLLM baseline on their corresponding inference modality (e.g., Unimodal MLLM achieves 74.57\% vs. 72.96\% for the Naive MLLM under egocentric inference). Despite being trained on additional modalities the naive MLLM fails to improve performance, suggesting that simply combining modalities without structured interaction does not result in effective knowledge transfer.

In contrast, MoP significantly improves performance across all inference modalities, achieving gains of \textcolor{codegreen}{+6.50\%}, \textcolor{codegreen}{+7.14\%}, and \textcolor{codegreen}{+11.84\%} over the Naive MLLM for egocentric, exocentric, and depth inference, respectively. Notably, MoP also outperforms the corresponding modality-only models for all three inference settings. Since these baselines are trained directly on the target inference modality alone, this improvement shows that MoP successfully leverages privileged modalities during training to improve single-modality inference. MoP also outperforms both X-InstructBLIP~\cite{panagopoulou2024xinstructblip} and OneLLM~\cite{han2024onellm} across all three inference modalities, showing that its gains are not simply due to stronger multimodal alignment, but to more effective use of privileged modalities.

\noindent\textbf{Music-AVQA.}
Table \ref{tab:sota_music-understanding} reports results on Music-AVQA under audio and video inference.
Unlike the ADL setting, naive multimodal training already improves over the unimodal baselines (53.44 vs. 46.75 under audio inference and 73.87 vs. 69.31 under video inference), indicating that paired audio-visual data is complementary even without structured interaction. MoP builds substantially on this, achieving further gains of \textcolor{codegreen}{+3.77} and \textcolor{codegreen}{+3.71} points over the Naive MLLM under audio and video inference, respectively.
Notably, MoP also outperforms recent alignment-based MLLMs, including X-InstructBLIP and OneLLM, under both audio and video inference. While these methods are trained with privileged modalities, their objective is primarily to align each modality with the LLM space, rather than to structure cross-modal representations for single-modality inference. The consistent gains of MoP suggest that alignment alone is not sufficient in the privileged modality setting; explicitly separating modality-specific and modality-general information provides a more effective mechanism for transferring knowledge from auxiliary modalities.

\begin{figure*}[h!]
    \centering
    \vspace{-0.25cm}
    \begin{minipage}[t]{0.32\textwidth}
        \centering
        \includegraphics[width=\linewidth]{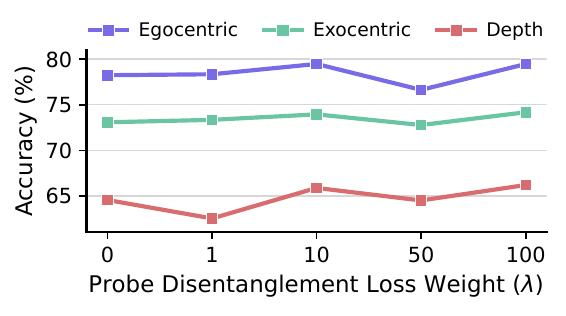}
        \vspace{-0.6cm}
        \captionof{figure}{\textbf{Ablation on probe disentanglement loss weight.}}
        \label{fig:ablation_div_loss_weight}
    \end{minipage}
    \hfill
    \begin{minipage}[t]{0.32\textwidth}
        \centering
        \includegraphics[width=\linewidth]{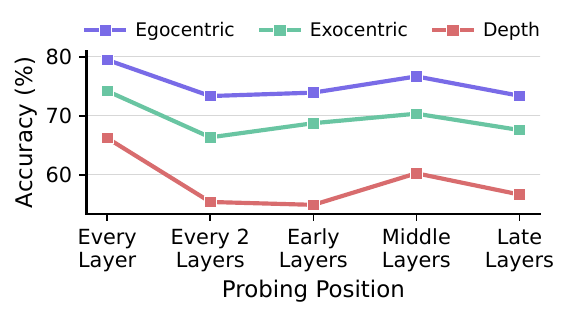}
        \vspace{-0.6cm}
        \captionof{figure}{\textbf{Ablation on position of probing layers.}}
        \label{fig:ablation_probe_locations}
    \end{minipage}
    \hfill
    \begin{minipage}[t]{0.32\textwidth}
        \centering
        \includegraphics[width=\linewidth]{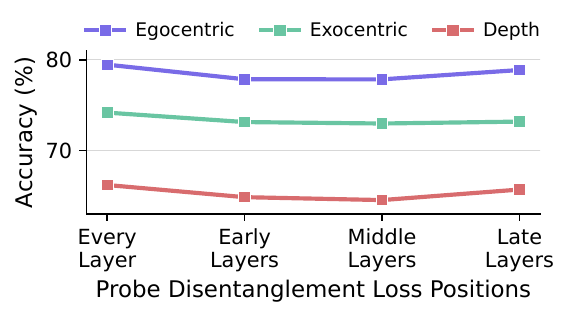}
        \vspace{-0.6cm}
        \captionof{figure}{\textbf{Ablation on application of disentanglement loss.}}
        \label{fig:ablation_div_loss_position}
    \end{minipage}
    \vspace{-0.3cm}
\end{figure*}

\vspace{-0.2cm}
\subsection{Ablation Study}
\vspace{-0.2cm}
We perform all ablations on the three inference settings of the Ego-in-Exo Perception benchmark. Additional ablations are presented in the Appendix.

\begin{wraptable}{o}{0.5\textwidth}
    \setlength{\tabcolsep}{6pt}
    \renewcommand{\arraystretch}{1.2}
    \centering
    \vspace{-0.25cm}
    \caption{\textbf{Ablation of Probe Types.} Impact of modality-specific and modality-general probing across different inference modalities.}
    \resizebox{\linewidth}{!}{
    \begin{tabular}{l|ccc}
         \hline
         \multirow{2}{*}{\textbf{Probing type}} &
         \multicolumn{3}{c}{\cellcolor{gray!20}\textbf{Inference Modality}} \\
         & \textbf{Egocentric} & \textbf{Exocentric} & \textbf{Depth} \\
         \hline
         Naive MLLM (No Probing) & 72.96 & 67.04 & 54.37 \\
         + Modality-specific & 76.72 (\textcolor{codegreen}{+3.76}) & 72.87 (\textcolor{codegreen}{+5.83}) & 63.22 (\textcolor{codegreen}{+8.85}) \\
         + Modality-general & \textbf{78.24 (\textcolor{codegreen}{+5.28})} & \textbf{73.08 (\textcolor{codegreen}{+6.04})} & \textbf{64.57 (\textcolor{codegreen}{+10.2})} \\
         \hline
    \end{tabular}}
    \vspace{-0.25cm}
    \label{tab:ablation_probe_type}
\end{wraptable}
\noindent\textbf{Effect of probing types.}
Table \ref{tab:ablation_probe_type} studies the contribution of the two probe families by comparing the naive MLLM, MoP with only modality-specific probes, and MoP with both modality-specific and modality-general probes. Adding modality-specific probes already yields clear gains across all three inference modalities, showing that intermediate-layer probing is effective even when restricted to modality-dependent information. Incorporating modality-general probes provides a further improvement in every case, leading to the best overall performance on egocentric, exocentric, and depth inference. These results support the core design of MoP: modality-specific probes capture information unique to each modality, while modality-general probes capture shared structure that is complementary.

\noindent\textbf{Effect of probe disentanglement loss.}
We study the effect of the probe disentanglement loss weight $\lambda$ in Figure \ref{fig:ablation_div_loss_weight} by varying its contribution relative to the autoregressive objective. Overall, MoP is reasonably stable across a broad range of $\lambda$, indicating that performance does not depend on highly sensitive tuning of the loss term. We observe that $\lambda = 100$ yields the strongest overall performance across all three inference modalities, and we therefore use this value in the remaining experiments.

\noindent\textbf{Location of probing layers.}
In Figure \ref{fig:ablation_probe_locations}, we ablate where probing is applied within the universal modality encoder. Specifically, we compare probing at every layer, every other layer, only early layers (1-5), only middle layers (5-15), and only late layers (15-27). Probing at every layer performs best overall, suggesting that MoP benefits from interacting with the encoder throughout its entire depth rather than only at a restricted subset of layers.

\noindent\textbf{Placement of probe disentanglement loss.}
In Figure \ref{fig:ablation_div_loss_position}, we study where the probe disentanglement loss is applied within the universal modality encoder. We compare applying the loss at every layer against restricting it to early, middle, or late layers only. Applying the probe disentanglement loss at every layer yields the strongest overall results, indicating that encouraging probe specialization throughout the full encoder is more effective than constraining it only locally.

\noindent\textbf{Controlled comparisons.}
In Table~\ref{tab:controlled}, we test whether MoP's added capacity (+3.7\% active parameters per sample and +5.65\% training compute) explains its gains by comparing against baselines that match or exceed its budget without the probing structure: an expanded projector, a Q-Former~\cite{blip2} on the final encoder layer, and a Q-Former fed by learnable tokens~\cite{darcet2024registers} traversing every encoder layer~\cite{jia2022visualprompttuning}. We further test explicit cross-modal distillation~\cite{gupta2016crossmodal} and unimodal training on data duplicated to match MoP's sample count. Every alternative trails MoP by at least 4.9 points on each inference modality, showing that the gains stem from the probe structure rather than from added parameters, compute, tokens, or data.

\begin{table*}[t]
    \setlength{\tabcolsep}{6pt}
    \renewcommand{\arraystretch}{1.2}
    \centering
    \vspace{-0.2cm}
    \caption{\textbf{Controlled comparisons on Ego-in-Exo Perception.} These baselines match or exceed MoP's parameters, compute, or training samples without its probe structure. Params column shows trainable parameters (total / active per sample).}
    \vspace{-0.2cm}
    \resizebox{0.96\linewidth}{!}{
    \begin{tabular}{l|cc|ccc}
         \hline
         \multirow{2}{*}{\textbf{Method}}&
         \multicolumn{2}{c|}{\cellcolor{gray!20}\textbf{Training Budget}} &
         \multicolumn{3}{c}{\cellcolor{gray!20}\textbf{Inference Modality}} \\
         & \textbf{Params (B)} & \textbf{TFLOPs ($\times10^6$)}
         & \textbf{Egocentric} & \textbf{Exocentric} & \textbf{Depth} \\
         \hline
         Naive MLLM~\cite{damonlpsg2025videollama3} & 0.178 / 0.178 & 71.3 & 72.96 & 67.04 & 54.37 \\
         \hline
         \multicolumn{6}{c}{\cellcolor{gray!20}\textit{Capacity-matched}} \\
         \hline
         Expanded Adapter & 0.487 / 0.487 & 73.4 & 72.78 & 67.41 & 59.10 \\
         Q-Former (final layer) & 0.767 / 0.767 & 75.5 & 73.10 & 66.77 & 55.08 \\
         Q-Former + Learnable Tokens & 0.767 / 0.767 & 75.8 & 72.42 & 66.09 & 53.52 \\
         \hline
         \multicolumn{6}{c}{\cellcolor{gray!20}\textit{Cross-modal distillation and data-matched}} \\
         \hline
         Cross-modal Distillation & 0.178 / 0.178 & 73.6 & 72.68 & 67.05 & 54.44 \\
         Unimodal MLLM (3$\times$ data) & 0.178 / 0.178 & 71.3 & 74.52 & 67.70 & 56.42 \\
         \hline
         \rowcolor{neuripspurple}\textbf{MoP (Ours)} & 0.769 / 0.482 & 75.3 & \textbf{79.46} & \textbf{74.18} & \textbf{66.21} \\
         \hline
    \end{tabular}}
    \vspace{-0.4cm}
    \label{tab:controlled}
\end{table*}


\begin{figure}[h]
    \centering
    \begin{subfigure}[t]{0.7425\linewidth}
        \centering
        \includegraphics[width=\linewidth]{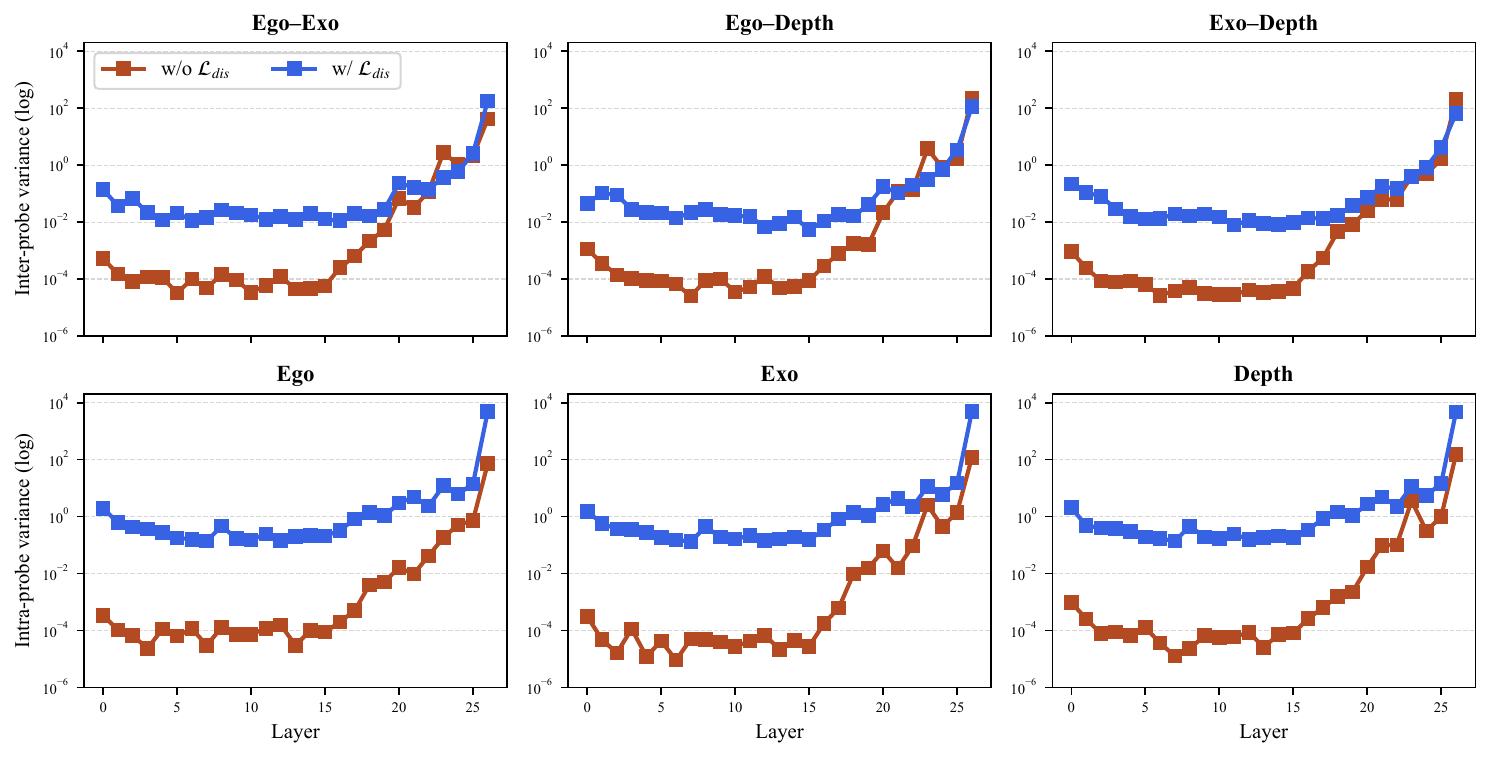}
        \vspace{-0.5cm}
        \caption{\textbf{Probe variance across universal modality encoder depth.}}
        \label{fig:analysis_probe_variance}
    \end{subfigure}
    \hfill
    \begin{subfigure}[t]{0.25\linewidth}
        \centering
        \includegraphics[width=\linewidth]{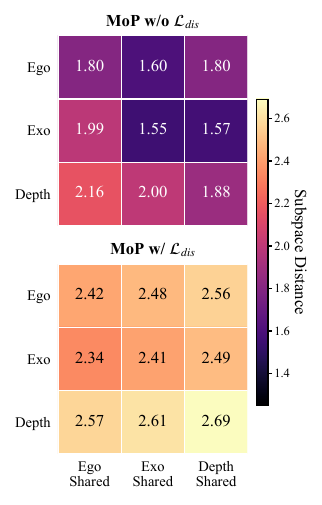}
        \vspace{-0.5cm}
        \caption{\textbf{Subspace distances.}}
        \label{fig:analysis_subspace_distances}
    \end{subfigure}
    
    \caption{\textbf{Effect of probe disentanglement loss on learned probe representations.}}
    \label{fig:analysis_disentanglement_effects}
    \vspace{-0.5cm}
\end{figure}

\vspace{-0.2cm}
\subsection{Model Diagnosis \& Analysis of MoP}
\vspace{-0.2cm}
\noindent\textbf{Effect of probe disentanglement loss on learned probes.}
Figure \ref{fig:analysis_probe_variance} analyzes the variance of the probe representations with and without probe disentanglement loss. The top row reports \emph{inter-probe variance} (across modality pairs), while the bottom row reports \emph{intra-probe variance} (within each modality). Without probe disentanglement loss, probe variance remains consistently low, indicating that probes collapse toward similar representations, particularly in early and mid-level layers. With the probe disentanglement loss, both inter- and intra-probe variance increase across layers and modalities, suggesting that probes learn more diverse and specialized representations.

To further understand this behavior, we analyze the geometry of the learned representations through pairwise subspace distances in Figure \ref{fig:analysis_subspace_distances}. At a high level, we ask whether modality-specific and shared probes occupy distinct directions in representation space. We measure this via the Frobenius distance between the SVD-extracted projection matrices of each probe group, where larger values indicate less directional overlap. Without the disentanglement loss, these distances tend to be smaller, suggesting the two probe groups orient similarly. With it, distances are consistently larger, consistent with specific probes being pushed toward more orthogonal structure relative to the shared subspace.

\begin{wrapfigure}{o}{0.46\textwidth}
    \centering
    \vspace{-0.4cm}
    \includegraphics[width=0.98\linewidth]{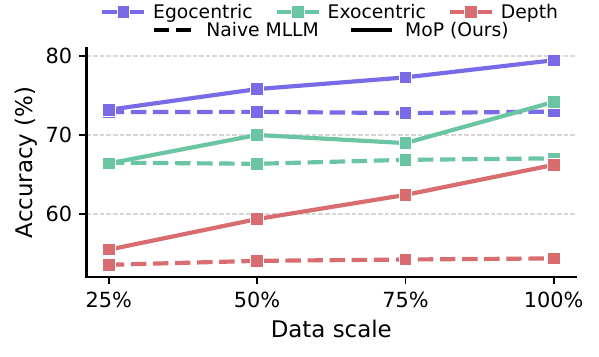}
    \vspace{-0.25cm}
    \caption{\textbf{Data scaling property of MoP.}}
    \label{fig:analysis_data_scaling}
\end{wrapfigure}
\noindent\textbf{MoP data scaling.}
Figure \ref{fig:analysis_data_scaling} studies how performance changes as the amount of training data increases from 25\% to 100\%. At each scale, both models are trained on the same subset, where each sample contains a single modality and its instruction, so the number of samples, tokens, and underlying events is matched. The naive MLLM baseline remains largely flat across all three inference modalities, indicating limited benefit from additional training data. In contrast, MoP improves consistently with scale, rising from 73.19 to 79.46 on egocentric inference, from 66.40 to 74.18 on exocentric inference, and from 55.48 to 66.21 on depth inference. These results suggest that MoP is able to exploit additional supervision more effectively than naive multimodal instruction tuning.

\begin{wraptable}{l}{0.54\textwidth}
    \setlength{\tabcolsep}{4pt}
    \renewcommand{\arraystretch}{1.2}
    \centering
    \vspace{-1em}
    \caption{\textbf{Computational overhead of MoP.}}
    \resizebox{\linewidth}{!}{
    \begin{tabular}{c|cc|cc}
         \hline
         \multirow{3}{*}{\textbf{Model}} &
         \multicolumn{2}{c|}{\cellcolor{gray!20}\textbf{Training}} &
         \multicolumn{2}{c}{\cellcolor{gray!20}\textbf{Inference}} \\
         & \textbf{Total Params} & \textbf{TFLOPs} & \textbf{Max VRAM} & \textbf{Latency} \\
         & \textbf{(All / Active)} & \textbf{($\times10^6$)} & \textbf{(GB)} & \textbf{(s)} \\
         \hline
         Naive MLLM & 8.21B / 8.21B & 71.3 & 24.4 & 0.767 \\
         MoP (Ours) & +7.2\% / +3.7\% & 75.3 (+5.65\%) & 32.5 & 1.078 \\
         \hline
    \end{tabular}}
    \label{tab:computation_overhead}
    \vspace{-1.25em}
\end{wraptable}

\noindent\textbf{Computational cost.}
In Table~\ref{tab:computation_overhead}, we report MoP's training and inference overhead relative to the Naive MLLM. Training compute increases by 5.65\%, and the layer-wise probe cross-attention raises inference latency from 0.767s to 1.078s per sample and peak VRAM from 24.4GB to 32.5GB. Because routing activates a single modality-specific branch per sample, MoP adds 7.2\% total parameters but only 3.7\% active parameters. MoP requires only one modality at inference, so its cost scales with the inference modality rather than the number of training modalities.



\section{Conclusion and Limitations}
\vspace{-0.2cm}
We introduce \textbf{Mixture of Probes (MoP)} for multimodal large language models under the \emph{privileged modality setting}.
MoP a probing-based framework that improves cross-modal knowledge transfer by separating modality-specific information from modality-general structure.
Our results demonstrate that this separation is the key to making auxiliary modalities useful: when properly leveraged, they can substantially improve single-modality performance. Despite these promising results, our study has two natural limitations. First, MoP assumes a universal modality encoder and does not directly extend to heterogeneous modality-specific encoders; this design is intentional, as the probing mechanism relies on a common representation space to extract and compare signals across modalities at intermediate layers. Second, we focus on the strict privileged modality setting with a single modality at inference and do not address multi-modal inference scenarios. This choice isolates the core challenge of transferring knowledge from auxiliary modalities. We leave extensions to heterogeneous encoders and multiple test-time modalities as important directions for future work.

\begin{ack}
\vspace{-0.2cm}
We would like to thank Toshimitsu Uesaka for carefully reading the manuscript and providing us with valuable feedback. This work was supported in part by the National Science Foundation (IIS-2245652) and the NSF National AI Research Resource Pilot (NAIRR240338).
\end{ack}

\newpage
\bibliographystyle{plain}
\bibliography{refs, llavidal, egoexo}

@String(CVPR= {IEEE Conf. Comput. Vis. Pattern Recog.})

@String(ICCV= {Int. Conf. Comput. Vis.})

@String(ECCV= {Eur. Conf. Comput. Vis.})

@String(ICPR = {Int. Conf. Pattern Recog.})

@String(ICLR = {Int. Conf. Learn. Represent.})

@String(AAAI = {AAAI})

@inproceedings{1,
  title     = {Learning realistic human actions from movies},
  author    = {Ivan Laptev and Marcin Marszałek and Cordelia Schmid and Benjamin Rozenfeld},
  booktitle = {CVPR},
  year      = {2008}
}

@inproceedings{2,
  title     = {Recognizing human actions: A local SVM approach},
  author    = {Christian Schuldt and Ivan Laptev and Barbara Caputo},
  booktitle = {ICPR},
  year      = {2004}
}

@inproceedings{3,
  title     = {Action recognition with improved trajectories},
  author    = {Heng Wang and Cordelia Schmid},
  booktitle = {ICCV},
  year      = {2013}
}

@inproceedings{5,
  title     = {Super Normal Vector for Activity Recognition Using Depth Sequences},
  author    = {Xiaodong Yang and Yingli Tian},
  booktitle = {CVPR},
  year      = {2014}
}

@inproceedings{7,
  title     = {Bilinear heterogeneous information machine for {RGB}-{D} action recognition},
  author    = {Kong, Yu and Fu, Yun},
  booktitle = {CVPR},
  year      = {2015}
}

@inproceedings{16,
  title     = {Articulated people detection and pose estimation:Reshaping the future},
  author    = {Leonid Pishchulin  and Arjun Jain and Mykhaylo Andriluka and Thorsten Thormahlen and Bernt Schiele},
  booktitle = {CVPR},
  year      = {2012}
}

@InProceedings{ego4d,
    author    = {Grauman, Kristen and Westbury, Andrew and Byrne, Eugene and Chavis, Zachary and Furnari, Antonino and Girdhar, Rohit and Hamburger, Jackson and Jiang, Hao and Liu, Miao and Liu, Xingyu and Martin, Miguel and Nagarajan, Tushar and Radosavovic, Ilija and Ramakrishnan, Santhosh Kumar and Ryan, Fiona and Sharma, Jayant and Wray, Michael and Xu, Mengmeng and Xu, Eric Zhongcong and Zhao, Chen and Bansal, Siddhant and Batra, Dhruv and Cartillier, Vincent and Crane, Sean and Do, Tien and Doulaty, Morrie and Erapalli, Akshay and Feichtenhofer, Christoph and Fragomeni, Adriano and Fu, Qichen and Gebreselasie, Abrham and Gonz\'alez, Cristina and Hillis, James and Huang, Xuhua and Huang, Yifei and Jia, Wenqi and Khoo, Weslie and Kol\'a\v{r}, J\'achym and Kottur, Satwik and Kumar, Anurag and Landini, Federico and Li, Chao and Li, Yanghao and Li, Zhenqiang and Mangalam, Karttikeya and Modhugu, Raghava and Munro, Jonathan and Murrell, Tullie and Nishiyasu, Takumi and Price, Will and Ruiz, Paola and Ramazanova, Merey and Sari, Leda and Somasundaram, Kiran and Southerland, Audrey and Sugano, Yusuke and Tao, Ruijie and Vo, Minh and Wang, Yuchen and Wu, Xindi and Yagi, Takuma and Zhao, Ziwei and Zhu, Yunyi and Arbel\'aez, Pablo and Crandall, David and Damen, Dima and Farinella, Giovanni Maria and Fuegen, Christian and Ghanem, Bernard and Ithapu, Vamsi Krishna and Jawahar, C. V. and Joo, Hanbyul and Kitani, Kris and Li, Haizhou and Newcombe, Richard and Oliva, Aude and Park, Hyun Soo and Rehg, James M. and Sato, Yoichi and Shi, Jianbo and Shou, Mike Zheng and Torralba, Antonio and Torresani, Lorenzo and Yan, Mingfei and Malik, Jitendra},
    title     = {Ego4D: Around the World in 3,000 Hours of Egocentric Video},
    booktitle = {Proceedings of the IEEE/CVF Conference on Computer Vision and Pattern Recognition (CVPR)},
    month     = {June},
    year      = {2022},
    pages     = {18995-19012}
}

@inproceedings{39,
  title     = {Finding action tubes},
  author    = {Georgia Gkloxari and Jitendra Malik},
  booktitle = {CVPR},
  year      = {2015}
}

@inproceedings{40,
  title     = {Unstructured Human Activity Detection from RGBD Images},
  author    = {Jaeyong Sung and Colin Ponce and Bart Selman and Ashutosh Saxena},
  booktitle = {ICRA},
  year      = {2012}
}

@inproceedings{videollava,
  title={Video-LLaVA: Learning United Visual Representation by Alignment Before Projection},
  author={Lin, Bin and Zhu, Bin and Ye, Yang and Ning, Munan and Jin, Peng and Yuan, Li},
  booktitle={Conference on Empirical Methods in Natural Language Processing (EMNLP)},
  year={2024}
}

@inproceedings{videollama,
  title={Video-LLaMA: An Instruction-tuned Audio-Visual Language Model for Video Understanding},
  author={Hang Zhang and Xin Li and Lidong Bing},
  journal={ArXiv},
  booktitle={Conference on Empirical Methods in Natural Language Processing (EMNLP)},
  year={2023}
}

@inproceedings{timechat,
  title     = {TimeChat: A Time-sensitive Multimodal Large Language Model for Long Video Understanding},
  author    = {Ren, Shuhuai and Yao, Linli and Li, Shicheng and Sun, Xu and Hou, Lu},
  booktitle = {Proceedings of the IEEE/CVF Conference on Computer Vision and Pattern Recognition},
  year      = {2024}
}

@misc{openai2024gpt4ocard,
      title={GPT-4o System Card}, 
      author={OpenAI},
      year={2024},
      eprint={2410.21276},
      archivePrefix={arXiv},
      primaryClass={cs.CL},
      url={https://arxiv.org/abs/2410.21276}, 
}

@inproceedings{egoexo,
  title     = {Ego-exo: Transferring Visual Representations from Third-person to First-person Videos},
  author    = {Li, Yanghao and Nagarajan, Tushar and Xiong, Bo and Grauman, Kristen},
  booktitle = {Computer Vision and Pattern Recognition (CVPR)},
  pages     = {6943--6953},
  year      = {2021}
}

@inproceedings{aytar2016soundnet,
  title     = {SoundNet: Learning Sound Representations from Unlabeled Video},
  author    = {Yusuf Aytar and Carl Vondrick and Antonio Torralba},
  booktitle = {Advances in Neural Information Processing Systems},
  year      = {2016}
}

@inproceedings{blip2,
  title={BLIP-2: Bootstrapping Language-Image Pre-training with Frozen Image Encoders and Large Language Models},
  author={Junnan Li and Dongxu Li and Silvio Savarese and Steven C. H. Hoi},
  booktitle={International Conference on Machine Learning},
  year={2023},
  url={https://api.semanticscholar.org/CorpusID:256390509}
}

@inproceedings{liu2023_llava,
      title={Visual Instruction Tuning}, 
      author={Liu, Haotian and Li, Chunyuan and Wu, Qingyang and Lee, Yong Jae},
      booktitle={Advances in Neural Information Processing Systems (NeurIPS)},
      year={2023},
}

@article{damonlpsg2025videollama3,
  title={VideoLLaMA 3: Frontier Multimodal Foundation Models for Image and Video Understanding},
  author={Zhang, Boqiang and Li, Kehan and Cheng, Zesen and Hu, Zhiqiang and Yuan, Yuqian and Chen, Guanzheng and Leng, Sicong and Jiang, Yuming and Zhang, Hang and Li, Xin and Jin, Peng and Zhang, Wenqi and Wang, Fan and Bing, Lidong and Zhao, Deli},
  journal={arXiv preprint arXiv:2501.13106},
  year={2025},
  url = {https://arxiv.org/abs/2501.13106}
}

@article{Qwen2.5-VL,
  title={Qwen2.5-VL Technical Report},
  author={Bai, Shuai and Chen, Keqin and Liu, Xuejing and Wang, Jialin and Ge, Wenbin and Song, Sibo and Dang, Kai and Wang, Peng and Wang, Shijie and Tang, Jun and Zhong, Humen and Zhu, Yuanzhi and Yang, Mingkun and Li, Zhaohai and Wan, Jianqiang and Wang, Pengfei and Ding, Wei and Fu, Zheren and Xu, Yiheng and Ye, Jiabo and Zhang, Xi and Xie, Tianbao and Cheng, Zesen and Zhang, Hang and Yang, Zhibo and Xu, Haiyang and Lin, Junyang},
  journal={arXiv preprint arXiv:2502.13923},
  year={2025}
}

@article{qwen2025qwen25technicalreport,
  title   = {Qwen2.5 Technical Report},
  author  = {{Qwen Team} and An Yang and Baosong Yang and Beichen Zhang and Binyuan Hui and Bo Zheng and Bowen Yu and Chengyuan Li and Dayiheng Liu and Fei Huang and Haoran Wei and Huan Lin and Jian Yang and Jianhong Tu and Jianwei Zhang and Jianxin Yang and Jiaxi Yang and Jingren Zhou and Junyang Lin and Kai Dang and Keming Lu and Keqin Bao and Kexin Yang and Le Yu and Mei Li and Mingfeng Xue and Pei Zhang and Qin Zhu and Rui Men and Runji Lin and Tianhao Li and Tianyi Tang and Tingyu Xia and Xingzhang Ren and Xuancheng Ren and Yang Fan and Yang Su and Yichang Zhang and Yu Wan and Yuqiong Liu and Zeyu Cui and Zhenru Zhang and Zihan Qiu},
  journal = {arXiv preprint arXiv:2412.15115},
  year    = {2025}
}

@inproceedings{zhai2023siglipv1,
      title={Sigmoid Loss for Language Image Pre-Training}, 
      author={Zhai, Xiaohua and Mustafa, Basil and Kolesnikov, Alexander and Beyer, Lucas},
      booktitle={Proceedings of the IEEE/CVF International Conference on Computer Vision},
      year={2023},
      pages={}, 
      organization={IEEE}
}

@inproceedings{reilly2026egoexo,
  title     = {From My View to Yours: Learning Egocentric Cues from Exocentric Video using Privileged Egocentric Supervision},
  author    = {Dominick Reilly and Manish Kumar Govind and Le Xue and Srijan Das},
  booktitle = {Proceedings of the European Conference on Computer Vision (ECCV)},
  year      = {2026}
}

@inproceedings{cvpr2025egoexo4d,
      title={Ego-Exo4D: Understanding Skilled Human Activity from First- and Third-Person Perspectives}, 
      author={Grauman, Kristen and Westbury, Andrew and Torresani, Lorenzo and Kitani, Kris and Malik, Jitendra and Afouras, Triantafyllos and Ashutosh, Kumar and Baiyya, Vijay and Bansal, Siddhant and Boote, Bikram and Byrne, Eugene and Chavis, Zach and Chen, Joya and Cheng, Feng and Chu, Fu-Jen and Crane, Sean and Dasgupta, Avijit and Dong, Jing and Escobar, Maria and Forigua, Cristhian and Gebreselasie, Abrham and Haresh, Sanjay and Huang, Jing and Islam, Md Mohaiminul and Jain, Suyog and Khirodkar, Rawal and Kukreja, Devansh and Liang, Kevin J and Liu, Jia-Wei and Majumder, Sagnik and Mao, Yongsen and Martin, Miguel and Mavroudi, Effrosyni and Nagarajan, Tushar and Ragusa, Francesco and Ramakrishnan, Santhosh Kumar and Seminara, Luigi and Somayazulu, Arjun and Song, Yale and Su, Shan and Xue, Zihui and Zhang, Edward and Zhang, Jinxu and Castillo, Angela and Chen, Changan and Fu, Xinzhu and Furuta, Ryosuke and Gonzalez, Cristina and Gupta, Prince and Hu, Jiabo and Huang, Yifei and Huang, Yiming and Khoo, Weslie and Kumar, Anush and Kuo, Robert and Lakhavani, Sach and Liu, Miao and Luo, Mi and Luo, Zhengyi and Meredith, Brighid and Miller, Austin and Oguntola, Oluwatumininu and Pan, Xiaqing and Peng, Penny and Pramanick, Shraman and Ramazanova, Merey and Ryan, Fiona and Shan, Wei and Somasundaram, Kiran and Song, Chenan and Southerland, Audrey and Tateno, Masatoshi and Wang, Huiyu and Wang, Yuchen and Yagi, Takuma and Yan, Mingfei and Yang, Xitong and Yu, Zecheng and Zha, Shengxin Cindy and Zhao, Chen and Zhao, Ziwei and Zhu, Zhifan and Zhuo, Jeff and Arbelaez, Pablo and Bertasius, Gedas and Crandall, David and Damen, Dima and Engel, Jakob and Farinella, Giovanni Maria and Furnari, Antonino and Ghanem, Bernard and Hoffman, Judy and Jawahar, C. V. and Newcombe, Richard and Park, Hyun Soo and Rehg, James M. and Sato, Yoichi and Savva, Manolis and Shi, Jianbo and Shou, Mike Zheng and Wray, Michael},
      booktitle={Proceedings of the IEEE/CVF Conference on Computer Vision and Pattern Recognition},
      year={2024}
}

@inproceedings{yang2024depthanythingv2,
      title={Depth Anything V2},
      author={Yang, Lihe and Kang, Bingyi and Huang, Zilong and Zhao, Zhen and Xu, Xiaogang and Feng, Jiashi and Zhao, Hengshuang},
      booktitle={Advances in Neural Information Processing Systems},
      year={2024}
}

@inproceedings{reilly2025llavidal,
  title        = {LLAVIDAL: A Large Language-Vision Model for Daily Activities of Living},
  author       = {Reilly, Dominick and Chakraborty, Rajatsubhra and Sinha, Arkaprava and Govind, Manish Kumar and Wang, Pu and Bremond, Francois and Xue, Le and Das, Srijan},
  booktitle    = {Proceedings of the IEEE/CVF Conference on Computer Vision and Pattern Recognition},
  year         = {2025}
}

@inproceedings{das2019toyotasmarthome,
  title        = {Toyota Smarthome: Real-World Activities of Daily Living},
  author       = {Das, Srijan and Dai, Rui and Koperski, Michal and Minciullo, Luca and Garattoni, Lorenzo and Bremond, Francois and Francesca, Gianpiero},
  booktitle    = {Proceedings of the IEEE/CVF International Conference on Computer Vision},
  year         = {2019},
  pages        = {833--842}
}

@article{dai2022toyotasmarthomeuntrimmed,
  title        = {Toyota Smarthome Untrimmed: Real-World Untrimmed Videos for Activity Detection},
  author       = {Dai, Rui and Das, Srijan and Sharma, Saurav and Minciullo, Luca and Garattoni, Lorenzo and Bremond, Francois and Francesca, Gianpiero},
  journal      = {IEEE Transactions on Pattern Analysis and Machine Intelligence},
  year         = {2022},
  doi          = {10.1109/TPAMI.2022.3169976}
}

@inproceedings{han2024onellm,
  title     = {OneLLM: One Framework to Align All Modalities with Language},
  author    = {Jiaming Han and Kaixiong Gong and Yiyuan Zhang and Jiaqi Wang and Kaipeng Zhang and Dahua Lin and Yu Qiao and Peng Gao and Xiangyu Yue},
  booktitle = {Proceedings of the IEEE/CVF Conference on Computer Vision and Pattern Recognition (CVPR)},
  year      = {2024}
}

@inproceedings{panagopoulou2024xinstructblip,
  title     = {X-InstructBLIP: A Framework for Aligning X-Modal Instruction-Aware Representations to LLMs and Emergent Cross-Modal Reasoning},
  author    = {Artemis Panagopoulou and Le Xue and Ning Yu and Junnan Li and Dongxu Li and Shafiq Joty and Ran Xu and Silvio Savarese and Caiming Xiong and Juan Carlos Niebles},
  booktitle = {Proceedings of the European Conference on Computer Vision (ECCV)},
  year      = {2024}
}

@inproceedings{xue2023modality,
  title     = {The Modality Focusing Hypothesis: Towards Understanding Crossmodal Knowledge Distillation},
  author    = {Zihui Xue and Zhengqi Gao and Sucheng Ren and Hang Zhao},
  booktitle = {International Conference on Learning Representations (ICLR)},
  year      = {2023}
}

@article{beyer2024paligemma,
  title   = {PaliGemma: A Versatile 3B VLM for Transfer},
  author  = {Lucas Beyer and Andreas Steiner and Andr{\'e} Susano Pinto and Alexander Kolesnikov and Xiao Wang and Daniel Salz and Maxim Neumann and Ibrahim Alabdulmohsin and Michael Tschannen and Emanuele Bugliarello and Thomas Unterthiner and Daniel Keysers and Skanda Koppula and Fangyu Liu and Adam Grycner and Alexey Gritsenko and Neil Houlsby and Manoj Kumar and Keran Rong and Julian Eisenschlos and Rishabh Kabra and Matthias Bauer and Matko Bo{\v{s}}njak and Xi Chen and Matthias Minderer and Paul Voigtlaender and Ioana Bica and Ivana Balazevic and Joan Puigcerver and Pinelopi Papalampidi and Olivier Henaff and Xi Xiong and Radu Soricut and Jeremiah Harmsen and Xiaohua Zhai},
  journal = {arXiv preprint arXiv:2407.07726},
  year    = {2024}
}

@inproceedings{dai2023instructblip,
  title     = {InstructBLIP: Towards General-Purpose Vision-Language Models with Instruction Tuning},
  author    = {Wenliang Dai and Junnan Li and Dongxu Li and Anthony Meng Huat Tiong and Junqi Zhao and Weisheng Wang and Boyang Li and Pascale Fung and Steven Hoi},
  booktitle = {Advances in Neural Information Processing Systems (NeurIPS)},
  year      = {2023}
}

@inproceedings{zhang2023videollama,
  title     = {Video-LLaMA: An Instruction-Tuned Audio-Visual Language Model for Video Understanding},
  author    = {Hang Zhang and Xin Li and Lidong Bing},
  booktitle = {Proceedings of the 2023 Conference on Empirical Methods in Natural Language Processing (EMNLP)},
  year      = {2023}
}

@inproceedings{maaz2024videochatgpt,
  title     = {Video-ChatGPT: Towards Detailed Video Understanding via Large Vision and Language Models},
  author    = {Muhammad Maaz and Hanoona Rasheed and Salman Khan and Fahad Shahbaz Khan},
  booktitle = {Proceedings of the 62nd Annual Meeting of the Association for Computational Linguistics (ACL)},
  year      = {2024}
}

@inproceedings{ye2024mplugowl2,
  title     = {mPLUG-Owl2: Revolutionizing Multi-modal Large Language Model with Modality Collaboration},
  author    = {Qinghao Ye and Haiyang Xu and Jiabo Ye and Ming Yan and Anwen Hu and Haowei Liu and Qi Qian and Ji Zhang and Fei Huang and Jingren Zhou},
  booktitle = {Proceedings of the IEEE/CVF Conference on Computer Vision and Pattern Recognition (CVPR)},
  year      = {2024}
}

@article{han2024imagebindllm,
  title   = {ImageBind-LLM: Multi-modality Instruction Tuning},
  author  = {Jiaming Han and Renrui Zhang and Wenqi Shao and Peng Gao and Peng Xu and Han Xiao and Kaipeng Zhang and Chris Liu and Song Wen and Ziyu Guo and Xudong Lu and Shuai Ren and Yafei Wen and Xiaoxin Chen and Xiangyu Yue and Hongsheng Li and Yu Qiao},
  journal = {arXiv preprint arXiv:2309.03905},
  year    = {2023}
}

@inproceedings{lu2022pretrainedtransformer,
  title     = {Pretrained Transformers as Universal Computation Engines},
  author    = {Kevin Lu and Aditya Grover and Pieter Abbeel and Igor Mordatch},
  booktitle = {Proceedings of the AAAI Conference on Artificial Intelligence (AAAI)},
  year      = {2022}
}

@article{zhang2023metatransformer,
  title   = {Meta-Transformer: A Unified Framework for Multimodal Learning},
  author  = {Yiyuan Zhang and Kaixiong Gong and Kaipeng Zhang and Hongsheng Li and Yu Qiao and Wanli Ouyang and Xiangyu Yue},
  journal = {arXiv preprint arXiv:2307.10802},
  year    = {2023}
}

@article{chen2023xllm,
  title   = {X-LLM: Bootstrapping Advanced Large Language Models by Treating Multi-Modalities as Foreign Languages},
  author  = {Feilong Chen and Minglun Han and Haozhi Zhao and Qingyang Zhang and Jing Shi and Shuang Xu and Bo Xu},
  journal = {arXiv preprint arXiv:2305.04160},
  year    = {2023}
}

@article{zhao2023chatbridge,
  title   = {ChatBridge: Bridging Modalities with Large Language Model as a Language Catalyst},
  author  = {Zijia Zhao and Longteng Guo and Tongtian Yue and Sihan Chen and Shuai Shao and Xinxin Zhu and Zehuan Yuan and Jing Liu},
  journal = {arXiv preprint arXiv:2305.16103},
  year    = {2023}
}

@inproceedings{moon2024anymal,
  title     = {AnyMAL: An Efficient and Scalable Any-Modality Augmented Language Model},
  author    = {Seungwhan Moon and Andrea Madotto and Zhaojiang Lin and Tushar Nagarajan and Matt Smith and Shashank Jain and Chun-Fu Yeh and Prakash Murugesan and Peyman Heidari and Yue Liu and Kavya Srinet and Babak Damavandi and Anuj Kumar},
  booktitle = {Proceedings of the 2024 Conference on Empirical Methods in Natural Language Processing (EMNLP)},
  year      = {2024}
}

@inproceedings{su2023pandagpt,
  title     = {PandaGPT: One Model To Instruction-Follow Them All},
  author    = {Yixuan Su and Tian Lan and Huayang Li and Jialu Xu and Yan Wang and Deng Cai},
  booktitle = {Proceedings of the 61st Annual Meeting of the Association for Computational Linguistics (ACL)},
  year      = {2023}
}

@inproceedings{li2022musicavqa,
  title={Learning to Answer Questions in Dynamic Audio-Visual Scenarios},
  author={Guangyao li and Yake Wei and Yapeng Tian and Chenliang Xu and Ji-Rong Wen and Di Hu},
  booktitle   = {IEEE Conference on Computer Vision and Pattern Recognition (CVPR)},
  year      = {2022},
}

@article{touvron2023llama,
  title   = {LLaMA: Open and Efficient Foundation Language Models},
  author  = {Hugo Touvron and Thibaut Lavril and Gautier Izacard and Xavier Martinet and Marie-Anne Lachaux and Timoth{\'e}e Lacroix and Baptiste Rozi{\`e}re and Naman Goyal and Eric Hambro and Faisal Azhar and Aurelien Rodriguez and Armand Joulin and Edouard Grave and Guillaume Lample},
  journal = {arXiv preprint arXiv:2302.13971},
  year    = {2023}
}

@misc{gemmateam2024gemma,
  title         = {Gemma: Open Models Based on Gemini Research and Technology},
  author        = {Gemma Team and Thomas Mesnard and Cassidy Hardin and Robert Dadashi and Surya Bhupatiraju and Shreya Pathak and Laurent Sifre and Morgane Rivi{\`e}re and Mihir Sanjay Kale and Juliette Love and Pouya Tafti and L{\'e}onard Hussenot and Pier Giuseppe Sessa and Aakanksha Chowdhery and Adam Roberts and Aditya Barua and Alex Botev and Alex Castro-Ros and Ambrose Slone and Am{\'e}lie H{\'e}liou and Andrea Tacchetti and Anna Bulanova and Antonia Paterson and Beth Tsai and Bobak Shahriari and Charline Le Lan and Christopher A. Choquette-Choo and Cl{\'e}ment Crepy and Daniel Cer and Daphne Ippolito and David Reid and Elena Buchatskaya and Eric Ni and Eric Noland and Geng Yan and George Tucker and George-Christian Muraru and Grigory Rozhdestvenskiy and Henryk Michalewski and Ian Tenney and Ivan Grishchenko and Jacob Austin and James Keeling and Jane Labanowski and Jean-Baptiste Lespiau and Jeff Stanway and Jenny Brennan and Jeremy Chen and Johan Ferret and Justin Chiu and Justin Mao-Jones and Katherine Lee and Kathy Yu and Katie Millican and Lars Lowe Sjoesund and Lisa Lee and Lucas Dixon and Machel Reid and Maciej Miku{\l}a and Mateo Wirth and Michael Sharman and Nikolai Chinaev and Nithum Thain and Olivier Bachem and Oscar Chang and Oscar Wahltinez and Paige Bailey and Paul Michel and Petko Yotov and Rahma Chaabouni and Ramona Comanescu and Reena Jana and Rohan Anil and Ross McIlroy and Ruibo Liu and Ryan Mullins and Samuel L Smith and Sebastian Borgeaud and Sertan Girgin and Sholto Douglas and Shree Pandya and Siamak Shakeri and Soham De and Ted Klimenko and Tom Hennigan and Vlad Feinberg and Wojciech Stokowiec and Yu-hui Chen and Zafarali Ahmed and Zhitao Gong and Tris Warkentin and Ludovic Peran and Minh Giang and Cl{\'e}ment Farabet and Oriol Vinyals and Jeff Dean and Koray Kavukcuoglu and Demis Hassabis and Zoubin Ghahramani and Douglas Eck and Joelle Barral and Fernando Pereira and Eli Collins and Armand Joulin and Noah Fiedel and Evan Senter and Alek Andreev and Kathleen Kenealy},
  year          = {2024},
  eprint        = {2403.08295},
  archivePrefix = {arXiv},
  primaryClass  = {cs.CL},
  url           = {https://arxiv.org/abs/2403.08295}
}

@inproceedings{brown2020gpt3,
  title     = {Language Models are Few-Shot Learners},
  author    = {Tom B. Brown and Benjamin Mann and Nick Ryder and Melanie Subbiah and Jared Kaplan and Prafulla Dhariwal and Arvind Neelakantan and Pranav Shyam and Girish Sastry and Amanda Askell and Sandhini Agarwal and Ariel Herbert-Voss and Gretchen Krueger and Tom Henighan and Rewon Child and Aditya Ramesh and Daniel M. Ziegler and Jeffrey Wu and Clemens Winter and Christopher Hesse and Mark Chen and Eric Sigler and Mateusz Litwin and Scott Gray and Benjamin Chess and Jack Clark and Christopher Berner and Sam McCandlish and Alec Radford and Ilya Sutskever and Dario Amodei},
  booktitle = {Advances in Neural Information Processing Systems (NeurIPS)},
  year      = {2020}
}

@inproceedings{jia2022visualprompttuning,
  title     = {Visual Prompt Tuning},
  author    = {Menglin Jia and Luming Tang and Bor-Chun Chen and Claire Cardie and Serge Belongie and Bharath Hariharan and Ser-Nam Lim},
  booktitle = {European Conference on Computer Vision},
  year      = {2022}
}

@inproceedings{raghu2021vitseelikecnn,
  title     = {Do Vision Transformers See Like Convolutional Neural Networks?},
  author    = {Maithra Raghu and Thomas Unterthiner and Simon Kornblith and Chiyuan Zhang and Alexey Dosovitskiy},
  booktitle = {Advances in Neural Information Processing Systems (NeurIPS)},
  year      = {2021}
}

@article{shotton2013kinect,
  title   = {Enhanced Computer Vision With Microsoft Kinect Sensor: A Review},
  author  = {Jungong Han and Ling Shao and Dong Xu and Jamie Shotton},
  journal = {IEEE Transactions on Cybernetics},
  volume  = {43},
  number  = {5},
  pages   = {1318--1334},
  year    = {2013},
  doi     = {10.1109/TCYB.2013.2265378}
}

@article{xu2025qwenomni25,
  title   = {Qwen2.5-Omni Technical Report},
  author  = {Jin Xu and Zhifang Guo and Jinzheng He and Hangrui Hu and Ting He and Shuai Bai and Keqin Chen and Jialin Wang and Yang Fan and Kai Dang and Bin Zhang and Xiong Wang and Yunfei Chu and Junyang Lin},
  journal = {arXiv preprint arXiv:2503.20215},
  year    = {2025}
}

@article{dang2024neuro_specific-general,
  title   = {Modality-specific and Modality-general Representations of Subjective Value in Frontal Cortex},
  author  = {Shilpa Dang and Jessica Emily Antono and Igor Kagan and Arezoo Pooresmaeili},
  journal = {Communications Biology},
  volume  = {7},
  pages   = {1550},
  year    = {2024}
}

@article{vapnik2009lupi,
  title   = {Learning Using Privileged Information: Similarity Control and Knowledge Transfer},
  author  = {Vladimir Vapnik and Rauf Izmailov},
  journal = {Journal of Machine Learning Research (JMLR)},
  year    = {2015}
}

@inproceedings{lopez2016generalized,
  title     = {Unifying Distillation and Privileged Information},
  author    = {David Lopez-Paz and L{\'e}on Bottou and Bernhard Sch{\"o}lkopf and Vladimir Vapnik},
  booktitle = {International Conference on Learning Representations (ICLR)},
  year      = {2016}
}

@inproceedings{gupta2016crossmodal,
  title     = {Cross Modal Distillation for Supervision Transfer},
  author    = {Saurabh Gupta and Judy Hoffman and Jitendra Malik},
  booktitle = {Proceedings of the IEEE Conference on Computer Vision and Pattern Recognition (CVPR)},
  year      = {2016}
}

@inproceedings{owens2016ambient,
  title     = {Ambient Sound Provides Supervision for Visual Learning},
  author    = {Andrew Owens and Jiajun Wu and Josh H. McDermott and William T. Freeman and Antonio Torralba},
  booktitle = {Proceedings of the European Conference on Computer Vision (ECCV)},
  year      = {2016}
}

@article{neverova2016moddrop,
  title   = {ModDrop: Adaptive Multi-Modal Gesture Recognition},
  author  = {Natalia Neverova and Christian Wolf and Graham W. Taylor and Florian Nebout},
  journal = {IEEE Transactions on Pattern Analysis and Machine Intelligence (TPAMI)},
  year    = {2016}
}

@inproceedings{ma2021missing,
  title     = {SMIL: Multimodal Learning with Severely Missing Modality},
  author    = {Mengmeng Ma and Jian Ren and Long Zhao and Sergey Tulyakov and Cathy Wu and Xi Peng},
  booktitle = {Proceedings of the AAAI Conference on Artificial Intelligence (AAAI)},
  year      = {2021}
}

@article{wu2026incomplete,
  title   = {Deep Multimodal Learning with Missing Modality: A Survey},
  author  = {Renjie Wu and Hu Wang and Hsiang-Ting Chen and Gustavo Carneiro},
  journal = {Transactions on Machine Learning Research (TMLR)},
  year    = {2026}
}

@inproceedings{pipoli2025missrag,
  title     = {MISSRAG: Addressing the Missing Modality Challenge in Multimodal Large Language Models},
  author    = {Vittorio Pipoli and Alessia Saporita and Federico Bolelli and Marcella Cornia and Lorenzo Baraldi and Costantino Grana and Rita Cucchiara and Elisa Ficarra},
  booktitle = {Proceedings of the IEEE/CVF International Conference on Computer Vision (ICCV)},
  year      = {2025}
}

@inproceedings{darcet2024registers,
  title={Vision Transformers Need Registers},
  author={Timothée Darcet and Maxime Oquab and Julien Mairal and Piotr Bojanowski},
  booktitle={International Conference on Learning Representations},
  year={2024}
}

\newpage
\appendix

\section{Broader impacts}
\vspace{-0.15cm}
The broader impacts of MoP stem from its ability to improve single-modality inference by leveraging privileged modalities available only during training, which can benefit real-world applications where high-fidelity sensors are useful for supervision but impractical, expensive, invasive, or unavailable at deployment, such as assistive robotics, healthcare monitoring, and content understanding. At the same time, because MoP is built on top of pretrained language models and multimodal encoders, it may inherit or amplify biases, hallucinations, privacy risks, and safety limitations associated with these foundation models. Moreover, since MoP transfers information from auxiliary modalities, care must be taken when those modalities contain sensitive information, such as egocentric video, audio, depth, or other sensor streams that may reveal personal behavior or private environments.

\section{Subspace distances (additional details)}
\vspace{-0.15cm}
To further understand how MoP affects the representation spaces of each modality, we analyze the geometry of the learned probe representations through pairwise subspace distances in Figure~\ref{fig:analysis_subspace_distances} on a subset of 500 samples of the Ego-in-Exo Perception benchmark. For a given probe group, we collect its learned probe representations into
\[
P \in \mathbb{R}^{N \times K \times d_v},
\]
where $N$ is the number of samples, $K$ is the number of probes in the group, and $d_v$ is the probe feature dimension. For modality-general probes, $K = K_g$, while for modality-specific probes, $K = K_s$. We then flatten this probe representation to obtain
$X \in \mathbb{R}^{(N K) \times d_v},$
so that each row corresponds to one probe representation. We center $X$ by subtracting its feature-wise mean, and then compute its top-$k$ right singular vectors using singular value decomposition:
\[
X = U \Sigma V^\top.
\]
The first $k$ right singular vectors define an orthonormal basis $B \in \mathbb{R}^{d_v \times k}$ for the dominant representation subspace of that probe group.

Given two probe groups with subspace bases $B_1$ and $B_2$, we compare their directions using the Frobenius distance between their projection matrices:
\[
d(B_1, B_2) = \left\| B_1 B_1^\top - B_2 B_2^\top \right\|_F.
\]
Smaller values indicate that two probe groups span similar directions in representation space, while larger values indicate less directional overlap. In our analysis, we use $k=10$ and compute this distance pairwise across probe groups.

\section{Additional Ablations}
\label{app:ablations}
\vspace{-0.7cm}
\begin{table}[h!]
    \setlength{\tabcolsep}{5pt}
    \renewcommand{\arraystretch}{1.2}
    \centering
    \begin{minipage}[t]{0.47\textwidth}
        \centering
        \caption{\textbf{Ablation on the number of probes.}}
        \resizebox{\linewidth}{!}{
        \begin{tabular}{cc|ccc}
             \hline
             \multirow{2}{*}{$\boldsymbol{K_s}$} & \multirow{2}{*}{$\boldsymbol{K_g}$} & \multicolumn{3}{c}{\cellcolor{gray!20}\textbf{Inference Modality}} \\
             & & \textbf{Egocentric} & \textbf{Exocentric} & \textbf{Depth} \\
             \hline
             4 & 4 & 77.27 & 71.03 & 65.27 \\
             8 & 4 & 78.74 & 71.52 & 64.77 \\
             4 & 8 & 78.48 & 73.56 & 64.00 \\
             \rowcolor{neuripspurple} 8 & 8 & \textbf{79.46} & \textbf{74.18} & \textbf{66.21} \\
             16 & 16 & 76.70 & 70.95 & 62.11 \\
             \hline
        \end{tabular}}
        \label{tab:probe_count}
    \end{minipage}\hfill
    \begin{minipage}[t]{0.5\textwidth}
        \centering
        \caption{\textbf{Ablation on privileged modality combinations.} -- denotes a modality absent from training.}
        \resizebox{\linewidth}{!}{
        \begin{tabular}{l|ccc}
             \hline
             \multirow{2}{*}{\textbf{Training Modalities}} & \multicolumn{3}{c}{\cellcolor{gray!20}\textbf{Inference Modality}} \\
             & \textbf{Egocentric} & \textbf{Exocentric} & \textbf{Depth} \\
             \hline
             Ego + Exo & 76.13 & 71.45 & -- \\
             Ego + Depth & 76.71 & -- & 62.65 \\
             Exo + Depth & -- & 70.15 & 60.93 \\
             \rowcolor{neuripspurple}\textbf{Ego + Exo + Depth} & \textbf{79.46} & \textbf{74.18} & \textbf{66.21} \\
             \hline
        \end{tabular}}
        \label{tab:modality_combos}
    \end{minipage}
\end{table}
\noindent\textbf{Number of probes.} Table~\ref{tab:probe_count} varies the number of modality-specific ($K_s$) and modality-general ($K_g$) probes. The 8/8 configuration performs best on all three inference modalities, and performance stays within about four points across configurations, so MoP does not depend on precise tuning of the probe count. Doubling to 16/16 yields the weakest results, further indicating that MoP's gains do not come from added probe capacity.

\noindent\textbf{Privileged modality combinations.} Table~\ref{tab:modality_combos} trains MoP on each pair of modalities, holding the architecture, recipe, and total number of training samples fixed. Each model is evaluated only on the modalities it was trained on, since a modality absent from training has no probes to route to. The full ego+exo+depth set outperforms every pairwise combination on each evaluated inference modality; under exocentric inference, for example, adding egocentric video to exo+depth training raises accuracy from 70.15 to 74.18. MoP's gains therefore accumulate across privileged modalities rather than being driven by any single one.

\begin{wrapfigure}{o}{0.52\textwidth}
    \centering
    \vspace{-0.4cm}
    \includegraphics[width=0.98\linewidth]{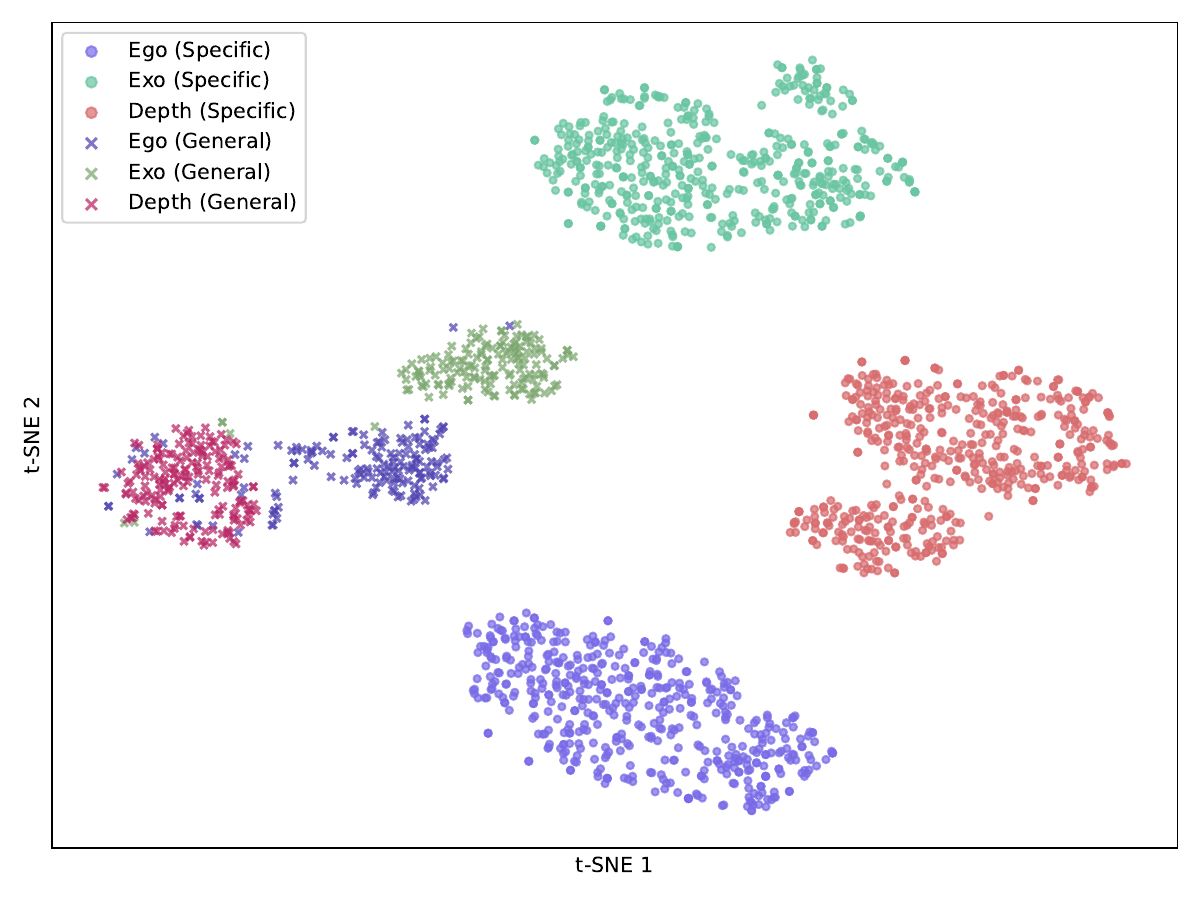}
    \vspace{-0.25cm}
    \caption{\textbf{t-SNE visualization of specific and general probes.}}
    \label{fig:appendix_tsne}
\end{wrapfigure}
\section{t-SNE analysis on learned probes}
\vspace{-0.15cm}
Figure~\ref{fig:appendix_tsne} visualizes the learned modality-specific and modality-general probe representations using t-SNE. We observe that the modality-specific probes form clearly separated clusters for egocentric, exocentric, and depth inputs, suggesting that they capture distinct modality-dependent structure. The modality-general probes form a cluster of their own, but we find that they additionally consist of smaller modality-conditioned sub-clusters. This behavior is consistent with the intended role of the two probe families: specific probes specialize to each modality, while general probes learn a shared representation space that is distinct from the specific probes. Importantly, the general probes do not collapse into identical representations across modalities, nor do they separate as strongly as the modality-specific probes, suggesting that they preserve transferable structure while retaining enough modality context to support cross-modal knowledge transfer.

\begin{figure}[h!]
    \centering
    \includegraphics[width=\linewidth]{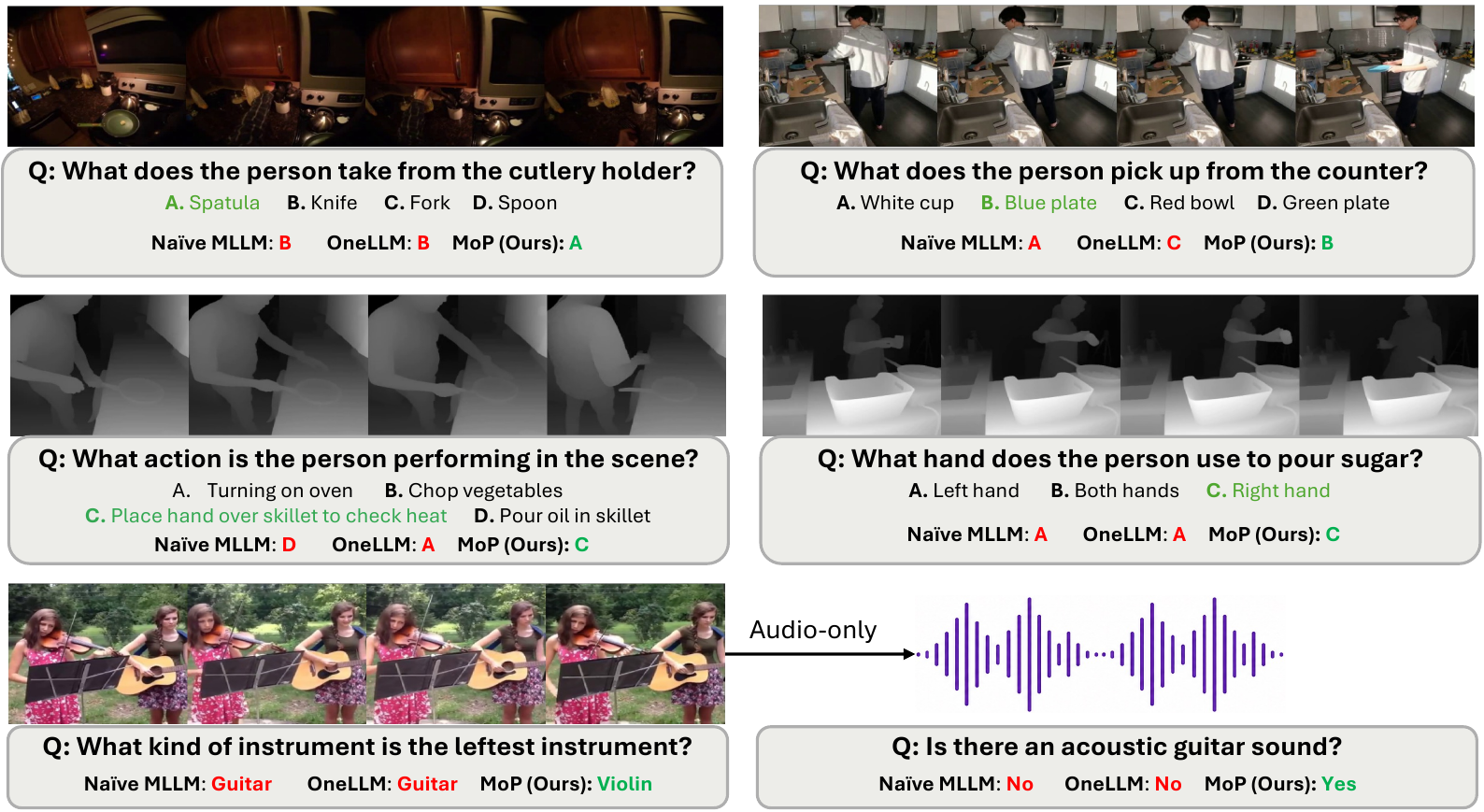}
    \caption{\textbf{Qualitative results of MoP on single modality inference.}}
    \label{fig:qualitative results}
\end{figure}
\section{Qualitative Results}
Figure~\ref{fig:qualitative results} presents qualitative comparisons between MoP and baseline MLLMs under single-modality inference. MoP produces more accurate responses, demonstrating a stronger ability to leverage the privileged modalities available during training. These examples qualitatively support the improvements observed in our quantitative evaluation.



\end{document}